\documentclass[conference]{IEEEtran}
\IEEEoverridecommandlockouts

\usepackage{amsmath,amsfonts,bm}

\usepackage{xcolor}
\usepackage{colortbl}

\def\eqref#1{equation~\ref{#1}}
\def\1{\bm{1}}

\DeclareMathAlphabet{\mathsfit}{\encodingdefault}{\sfdefault}{m}{sl}
\SetMathAlphabet{\mathsfit}{bold}{\encodingdefault}{\sfdefault}{bx}{n}

\newcommand{\E}{\mathbb{E}}

\DeclareMathOperator{\sign}{sign}

\newcommand\Ga{}

\newcommand\Gc{\rowcolor{orange!40}}

\usepackage[colorlinks=true,
citecolor=blue,
filecolor=black,
linkcolor=blue,
urlcolor=blue]{hyperref}
\usepackage{url}
\usepackage{graphicx}
\usepackage{multirow}
\usepackage{booktabs}
\usepackage{paralist}

\usepackage{booktabs} % for professional tables
\usepackage{multicol}

\usepackage[ruled,noend]{algorithm2e}

\SetCommentSty{mycommfont}

\usepackage{here}

\usepackage{graphicx}
\usepackage{caption}
\usepackage{amsmath,amssymb,amsfonts,amsbsy,amsfonts,latexsym}
\usepackage{multirow}
\usepackage{makecell}
\usepackage[labelfont=bf,textfont=it,belowskip=0pt,aboveskip=5pt,tableposition=top]{caption}
\usepackage{xcolor}
\usepackage{colortbl}

\definecolor{colorA}{RGB}{189,201,225}
\definecolor{colorB}{RGB}{103,169,207}
\definecolor{colorC}{RGB}{ 28,144,153}
\definecolor{colorD}{RGB}{  1,108, 89}

\newcolumntype{R}{>{\columncolor{gray!40}}r}
\newcolumntype{L}{>{\columncolor{gray!40}}l}
\newcolumntype{C}{>{\columncolor{gray!40}}c}

\usepackage{tabularx,colortbl,xcolor}
\usepackage{multirow}
\usepackage[normalem]{ulem}
\useunder{\uline}{\ul}{}

\usepackage{enumitem}

\usepackage{xparse}% http://ctan.org/pkg/xparse

\graphicspath{{fig}}
\DeclareGraphicsExtensions{.pdf,.png}
\usepackage{pgfplots, pgfplotstable}

\SetKwInput{KwInput}{Input}
\newcommand{\pluseq}{\mathrel{+}=}
\newcommand{\asteq}{\mathrel{*}=}
\newcommand{\diveq}{\mathrel{/}=}

\usepackage{tikz, pgfplots}

\usetikzlibrary{patterns}

\NewDocumentCommand{\var}{O{s} m O{}}{%
  \ensuremath{#1_{#2}^{#3}}% add \vphantom{<bizarre sup>}
}
\usepackage{siunitx}

\definecolor{light-gray}{gray}{0.80}

\renewcommand\paragraph{\subsubsection*}

\newcommand\aref{Alg.~\ref}

\newcommand\fref{Fig.~\ref}
\newcommand\tref{Tab.~\ref}

\newcommand\ha{ \rowcolor{orange!0}}

\newcommand\hc{ \rowcolor{orange!40}}

\def\0{{\bf 0}}

\def\V{{\mathbb{V}}}

\usepackage{amsthm}

\newtheorem{mytheorem}{Theorem}
\newtheorem{assumption}[mytheorem]{Assumption}
\newtheorem{lemma}[mytheorem]{Lemma}
\newtheorem{remk}[mytheorem]{Remark}

\graphicspath{{figures/}}
\usepackage[square,sort&compress,comma,numbers]{natbib}

\begin{document}

\title{\huge Large Batch Size Training of Neural Networks with Adversarial Training and Second-Order Information}
\author{\large
Zhewei Yao$^{*}$\thanks{$^{*}$Equal contribution.}, Amir Gholami$^{*}$, Daiyaan Arfeen, Richard Liaw,\\ Joseph Gonzalez, Kurt Keutzer, Michael W. Mahoney\\
University of California, Berkeley\\
{\normalsize \{zheweiy, amirgh, daiyaanarfeen, rliaw, jegonzal, keutzer, and mahoneymw\}@berkeley.edu}
}

\maketitle

The most straightforward method to accelerate Stochastic Gradient Descent (SGD)
computation is to distribute the randomly selected batch of inputs over multiple processors. 
To keep the distributed processors fully utilized requires commensurately growing the batch size. 
However, large batch training often leads to poorer generalization.
A recently proposed solution for this problem is to use adaptive batch sizes in SGD.
In this case, one starts with a small number of processes and scales the processes as training progresses.
Two major challenges with this approach are (i) that dynamically resizing the cluster can add non-trivial overhead, in part since it is currently not supported, and
(ii) that the overall speed up is limited by the initial phase with smaller batches.
In this work, we address both challenges by developing a new adaptive
batch size framework, with autoscaling based on the Ray framework.  This allows very efficient elastic scaling
with negligible resizing overhead (0.32\% of time for ResNet18 ImageNet training).
Furthermore, we propose a new adaptive batch size training scheme using second order methods and adversarial training.
These enable increasing batch sizes earlier during
training, which leads to better training time. 
We extensively evaluate our method on Cifar-10/100, SVHN, TinyImageNet, and ImageNet datasets, using multiple neural networks, including ResNets and smaller networks such as SqueezeNext. 
Our new approach exceeds the performance of existing solutions in terms of both accuracy and the number of SGD iterations (up to 1\% and $5\times$, respectively).
%We emphasize that 
Importantly,
this is achieved without any additional hyper-parameter tuning to tailor our proposed method in any of these experiments. With slight hyper-parameter tuning, our method can reduce the number of SGD iterations of ResNet20/18 on Cifar-10/ImageNet to $44.8\times$ and $28.8\times$,~respectively.

\section{Introduction}

%%%%%%%%%%%%%%%%%%%%%%%%%%%%%%%%%%%%%%%%%%%%%%%%%%%%%%%%%%%%%%%%%%%%%%%%%%%%%%%%%%%%%%
\begin{figure*}[tbp]
\begin{center}
  \includegraphics[width=.45\textwidth,trim={2.5cm 0 1.5cm 0},clip]
  {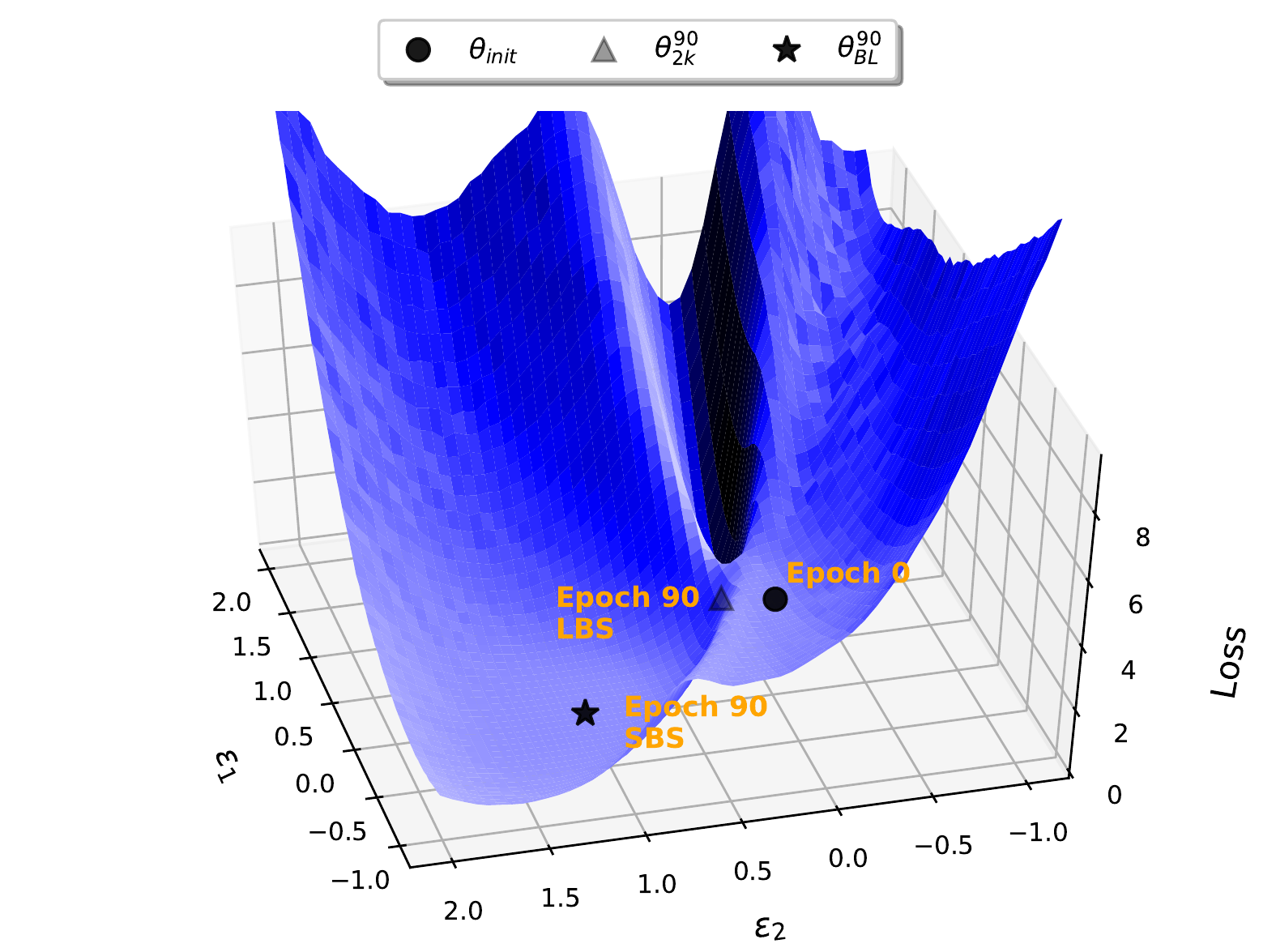}
    \includegraphics[width=.45\textwidth,trim={2.5cm 0 1.5cm 0},clip]{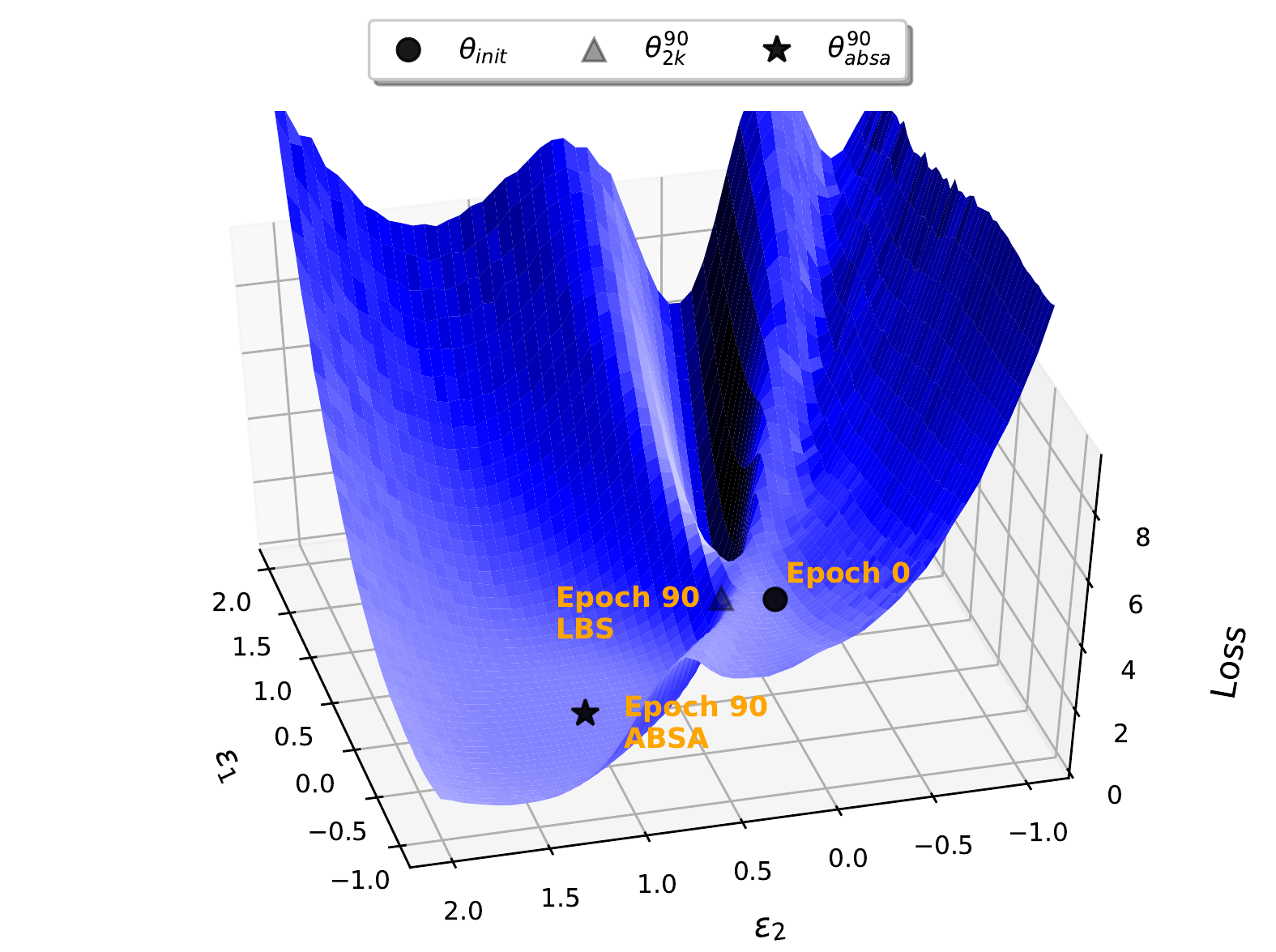}
\end{center}
\caption{
\footnotesize 
(Left) 3D parametric plot for C1 model on Cifar-10.
Points are labeled with the number of epochs (e.g., 90) and the technique that was used to arrive at that point (e.g., large batch size, LBS).
The $\epsilon_1$ direction shows
how the loss changes across the path between initial model at epoch 0, and the final model
achieved with Large Batch Size (LBS) of $B=2K$. Similarly, the $\epsilon_2$ direction computes
the loss when the model parameters are interpolated between epoch 0 and final model at epoch 90
with Small Batch Size (SBS). Note the sharp curvature directions to which LBS gets attracted. 
(Right)
A similar plot, except that we use the ABSA algorithm with final batch of $16K$, rather than SGD with a small batch size for interpolating the $\epsilon_2$ direction. 
Note the visual similarity between the point to which ABSA converges after 90 epochs (ABSA, 84.24\% accuracy) and the point to which small batch SGD (SBS, 83.05\% accuracy) converges after 90 epochs. Note also that both avoid the sharp landscape to which large batch gets attracted (LBS, 76.82\%). Generalization errors are shown in~\tref{tab:abs_cifar10_resnet18}.
}
\label{fig:2d_cifar_resnet}
\end{figure*}
%%%%%%%%%%%%%%%%%%%%%%%%%%%%%%%%%%%%%%%%%%%%%%%%%%%%%%%%%%%%%%%%%%%%%%%%%%%%%%%%%%%%%%

Finding the right neural network (NN) model architecture for a particular application typically requires extensive
hyper-parameter tuning and architecture search, often on a very large dataset. 
These delays associated with training NNs are routinely the main bottleneck in the design process. 
One way to address this issue is to use large distributed processor clusters to perform distributed Stochastic Gradient Descent (SGD). 
However, to efficiently utilize each 
processor, the portion of the batch associated with each processor (sometimes called the mini-batch size) must grow correspondingly. 
In the ideal case, the goal is to decrease the computational time proportional to the
increase in batch size, without any drop in generalization quality. 
However, large batch training has a number of well known 
drawbacks~\cite{golmant2018computational,mccandlish2018empirical,shallue2018measuring}.
These include degradation of accuracy, poor generalization, wasted computation, and even poor robustness to adversarial perturbations~\citep{keskar2016large,yao2018hessian}.

In order to address these drawbacks, many solutions have been proposed~\citep{goyal2017accurate, you2017scaling, devarakonda2017adabatch, smith2017don,jia2018highly}. 
However, these methods either work only for particular models on a particular dataset, or they require extensive hyper-parameter tuning~\cite{ma2019inefficiency,golmant2018computational,shallue2018measuring}. The latter point is often not discussed in the presentation of record-breaking training-times with large batch sizes.
However, these cost are important for a practitioner who wants to accelerate training time
with large batches to develop and use a model.

One recent solution is to incorporate adaptive batch size training~\citep{smith2017don,devarakonda2017adabatch}. The proposed method involves a hybrid increase of batch size and learning rate to accelerate training.
In
this approach, one would select
a strategy to ``anneal'' the batch size during training. This is based on
the idea that large batches
contain less ``noise,'' and that could be used in much the same way as reducing the learning rate during training or the temperature parameter in simulated annealing optimization. 

As the use of adaptive batch sizes results in the elastic use of distributed computing resources, this approach is ideally suited to the emerging 
``pay as you use'' 
era of Serverless Computing~\cite{baldini2017serverless}.  
However, there are two major limitations with this approach.
%(i) 
First, 
there is no framework that supports changing of batch size during training and rescaling of the processes.
Currently the only approach is to run the problem with a fixed batch size, snapshot the training, and restart a new
job with more processes.
%; and
%(ii)
Second, the training is started with small batches,
which can only be scaled to a small number of processes.
A simple application of Amdahl's law shows that the speed up bottleneck will be the portion of the training that uses
small batch since it cannot be parallelized to large number of processes.

In this paper, we develop methods to address both of these shortcomings. 
In more detail, our main contributions are the following.

\begin{itemize}[noitemsep,topsep=0pt,parsep=0pt,partopsep=0pt,leftmargin=*]

\item 
We propose an Adaptive Batch Size (ABS) method for SGD training that is based on second order information (i.e., Hessian information).
Our method automatically changes the batch size and learning rate based on the loss landscape curvature, as measured via the Hessian spectrum.
% We show a basic result that this method is convergent for a convex problem.
  %When the top eigenvalue w.r.t. weight decreases by a factor of $\alpha$,
  %we increase the batch size and learning rate by a factor of $\beta$.
% We empirically test the algorithm for important non-convex problems
% in deep learning and show that it achieves equal or better test performance, as compared to small batch SGD.

\item  
We propose an Adaptive Batch Size Adversarial (ABSA) method for SGD training.
This method extends ABS by using an implicit regularization approach based on robust training by solving a min-max optimization problem.
In particular, we combine the second order ABS method with recent results of~\citep{yao2018hessian} which show that robust training can be used to regularize against sharp minima.
We show that this combination of Hessian-based adaptive batch size and robust optimization achieves
significantly better test performance (up to 1\%; see Table~\ref{tab:abs_cifar10_wresnet} in Appendix) with little computational~overhead.
% \zhewei{need to quantify this, how much better?}
  
\item  
We extensively test our proposed solution on a wide range of datasets (Cifar-10/100, SVHN, TinyImageNet, and ImageNet), using different NNs, including residual networks. 
Importantly, we use the \emph{same hyper-parameters} for all of the experiments, and we do \emph{not} perform any kind of tuning to tailor our results. 
The empirical results show the clear benefit of our proposed method, as compared to the state-of-the-art.
The proposed algorithm achieves equal or better test accuracy (up to 1\%) and requires significantly fewer SGD updates (up to $5\times$). 
In addition, if we permit ourselves to perform a slight hyper-parameter tuning of the warmup schedule, then we show that the number of SGD updates can be reduced $44.8\times$ on Cifar-10 and $28.8\times$ on ImageNet.

\item
We develop a scalable adaptive batch size framework based on Ray~\cite{moritz2018ray}.
Our framework supports elastic resizing of the cluster to increase or decrease the number of processes.  It also supports the adversarial / second order based methods proposed above as well as the adaptive batch size method of~\cite{smith2017don}.
We present the scaling of our proposed method on Amazon Web Services for training
ResNet18 on ImageNet, showing negligible cluster resizing cost (0.32\% out of entire training time).

\end{itemize}

While a number of recent works have discussed adaptive batch size or 
increasing batch size during 
training~\citep{devarakonda2017adabatch,smith2017don,friedlander2012hybrid,balles2016coupling}, to the best of our knowledge this is the first paper to introduce a scalable framework that automatically resizes the batch size and cluster. Moreover, to the best of our knowledge, this is the first work that introduces Hessian information and adversarial training in adaptive batch size training, with extensive testing on many datasets.

% \vspace{-2mm}
\paragraph{Limitations}

We believe that it is important for every work to state its limitations (in 
general, but in particular in this area). 
We were particularly careful to perform extensive experiments, and we repeated 
all the reported tests multiple times.
We test the algorithm on models ranging from a few layers to hundreds of layers, including
residual networks as well as smaller networks such as SqueezeNext. We have also
open sourced our code to allow reproducibility~\cite{hessianflow}.
An important limitation is that second order methods have additional overhead
for backpropagating the Hessian. 
Currently, most of the existing frameworks do not support (memory) efficient 
backpropagation of the Hessian (thus providing a structural bias against 
these powerful methods). 
However, the complexity of each Hessian matvec is the same as that for a gradient 
computation~\citep{yao2019pyhessian,martens2010deep}.
Our method requires an estimate of the Hessian spectrum, which typically needs ten Hessian 
matvecs (for power method iterations to reach a relative tolerance of 1e-2).
Thus, while the additional cost is not too large, the benefits that we show in terms of testing accuracy and reduced 
number of updates do come at additional cost (see section~\ref{sec:time_measurement} for details).
We support this by showing measured Hessian overhead time for an actual training problem as shown in~\fref{fig:scalingbar}.

\section{Related Work}
\label{sec:related_work}
Optimization methods based on SGD are currently
the most effective techniques for training NNs, and this is commonly attributed to SGD's ability to escape saddle-points and ``bad'' local minima~\citep{dauphin2014identifying}.
However, the sequential nature of weight updates in synchronous SGD limits possibilities for parallel computing. 
In recent years, there has been considerable effort on breaking this sequential nature, through asynchronous methods~\citep{zhang2015deep} or symbolic execution techniques~\citep{maleki2017parallel}.
A main problem with asynchronous methods is reproducibility, which, in this case, depends on the number of processes used~\citep{zheng2016asynchronous,agarwal2011distributed}.
Due to this issue, recently there have been attempts to increase parallelization opportunities in synchronous SGD by using large batch size training. 
With large batches, it is possible to distribute computations more 
efficiently to parallel compute nodes~\citep{gholami2017integrated}, thus reducing the total training time.
However, large batch training often leads to sub-optimal test performance~\citep{keskar2016large,yao2018hessian}.
This has been attributed to the observation that large batch size training tends to get attracted to local minima or sharp curvature directions, which are not robust to (possible) mismatch between training and testing curves~\citep{keskar2016large}.
A full understanding of this, however, remains elusive (but recent work has used random matrix theory to characterize models trained with smaller or larger batch sizes in terms of stronger or weaker implicit regularization~\citep{MM18_TR,MM19_HTSR_ICML}).

\begin{algorithm}
\small
\DontPrintSemicolon
\caption{Adaptive Batch Size (ABS) and Adaptive Batch Size Adversarial (ABSA)}\label{alg:abs}
   \SetAlgoLined
    \KwInput{Learning rate $\eta$, decay epoch $E_{decay}$, decay ratio $\rho_{lr, decay}$; initial batch $B_{init}$, maximum batch $B_{max}$; Eigenvalue decay ratio $\alpha$, batch/lr increasing ratio $\beta$, duration factor $\kappa$; adversarial input ratio $\gamma$, adversarial magnitude $\epsilon_{adv}$, adversarial input ratio decay factor $\omega$, terminate epoch $\tau$; input data $x$. }
    \KwResult{Final model parameters $\theta$}
    Initialize old (new) eigenvalue as $\lambda_{old}$ ($\lambda_{new}$), batch size $B_{itr}$, learning rate $\eta_{itr}$, duration time $\kappa_{itr}=0$.
    
    \For{Epoch $=1,2,3\ldots$}{
      \For(\tcp*[h]{mini-batch iterations}){$t=1,2,3\ldots$}{
        generate adversarial data $x_{adv}$ by the ratio of $\gamma$;
        
        SGD update;
      }
     
      compute $\lambda_{new}$; \tcp*[h]{Hessian power-iteration}
      
      $\kappa_{itr} \pluseq 1$;
      
      $Flag = \{(\lambda_{new} < \lambda_{old}/\alpha)~OR~(\kappa_{itr}=\kappa)\}$;
       
      \If(\tcp*[h]{adaptively change batch/lr}){Flag}{
      $B_{itr}\asteq\beta$;~~
      $\eta_{itr}\asteq\beta$;
      $\gamma \diveq \omega$;~
      $\kappa_{itr}=0$;~
      }
      \If{$\lambda_{new} < \lambda_{old}/\alpha$}{
      $\lambda_{old}=\lambda_{new}$;
      }
      
      \If{Epoch $=\tau$ }{
        $\gamma = 0$;
      }
      \If(\tcp*[h]{lr decay schedule}){Epoch in $E_{decay}$}{
      $\eta_{itr} \diveq \rho_{lr, decay}$ ;
      }    
    }
\end{algorithm}

There have been several solutions proposed for alleviating the problem with 
large batch size training.
The first notable work here is~\citep{goyal2017accurate}, where it was shown 
that by scaling the learning rate,
it is possible to achieve the same testing accuracy for large batches. In 
particular, ResNet50 model
was tested on ImageNet dataset, and it was shown that the baseline accuracy 
could be recovered up to batch size
of 8192. However, this approach does not generalize to other networks/tasks~\citep{golmant2018computational}. In~\citep{you2017scaling}, an adaptive learning rate 
method %(called LARS) 
was proposed which allowed
scaling training to a much larger batch size of 32K with more hyper-parameter 
tuning.
More recent work \citep{jia2018highly} proposed mix-precision 
method to explore further  
the limit of large batch training.
There is also a set of work on dynamic batch size selection
to accelerate optimization, where the batch size is selected to ensure a search direction that would decrease the loss function with high probability~\cite{bollapragada2018progressive,byrd2012sample,bollapragada2018adaptive}.
Another notable work is the use of distributed K-FAC method to perform large batch training~\cite{osawa2018second}.

Work has also focused on using second order methods to reduce the brittleness of SGD. 
% Full Newton method with line search is parameter-free, and it does not require a learning rate.
% This is achieved by using a second-order Taylor series approximation to the loss function, instead of a first-order one as in SGD, to obtain curvature information.
Most notably, \citep{schaul2013no,xu2017second} show that Newton/quasi-Newton methods outperform SGD for training NNs. 
However, their results only consider simple fully-connected NNs and auto-encoders. 
%\citet{chen2018comparison} make a more extensive comparison between Newton/quasi-Newton methods and SGD, and they conclude that second order methods are not efficient, due to their slow computation.  
A problem with naive second-order methods is that they can exacerbate the large batch problem, as by construction they have a higher tendency to get attracted to local minima, as compared to SGD. 
For these reasons, early attempts at using second-order methods for training convolutional NNs have so far not been~successful.

A recent study has shown that anisotropic noise injection based on second order information could also help in 
escaping sharp landscape~\citep{zhu2018anisotropic}.
The authors showed that the noise from SGD could be viewed as anisotropic, 
with the Hessian as its covariance matrix.
Injecting random noise using the Hessian as covariance was proposed as a 
method to avoid sharp landscape points.

Another recent work by~\citep{yao2018hessian} has shown that adversarial 
training (or robust optimization)
could be used to ``regularize'' against these sharp minima,
with preliminary results showing superior testing performance, as compared
to other methods. 
In addition, \citep{shaham2015understanding,shrivastava2017learning} used adversarial training and showed that
the model trained using robust optimization is often more robust
to perturbations, as compared to normal SGD training. Similar observations
have been made by others~\citep{goodfellow6572explaining}.
These works are in line with recent results showing that adversarial attacks can sometimes be interpreted
as providing useful features~\cite{ilyas2019adversarial}.

Our results are in agreement with recent studies showing that there is a limit beyond which increasing batch size
will lead to diminishing results~\cite{golmant2018computational,ma2019inefficiency,shallue2018measuring}.
These studies mainly focus on constant batch size. However, the work of 
~\cite{golmant2018computational} showed that warmup with smaller batches can noticeably delay
this limit to larger batches. 
An adaptive batch size strategy would involve several different batches throughout the training, and  this adaptive schedule would be more efficient in the
regime of large batches~\cite{mccandlish2018empirical}. This means that increasing the batch size in each segment of this adaptive schedule would be more efficient in terms of 
minimum number of optimization steps~\cite{mccandlish2018empirical}. Our goal is to achieve this using ABS/ABSA algorithms to fully exploit larger batches during training,
and delay the limit beyond which using large batches would result in diminishing returns.

% -----------------------
\begin{figure}[!htbp]
\begin{center}
\includegraphics[width=.45\textwidth]{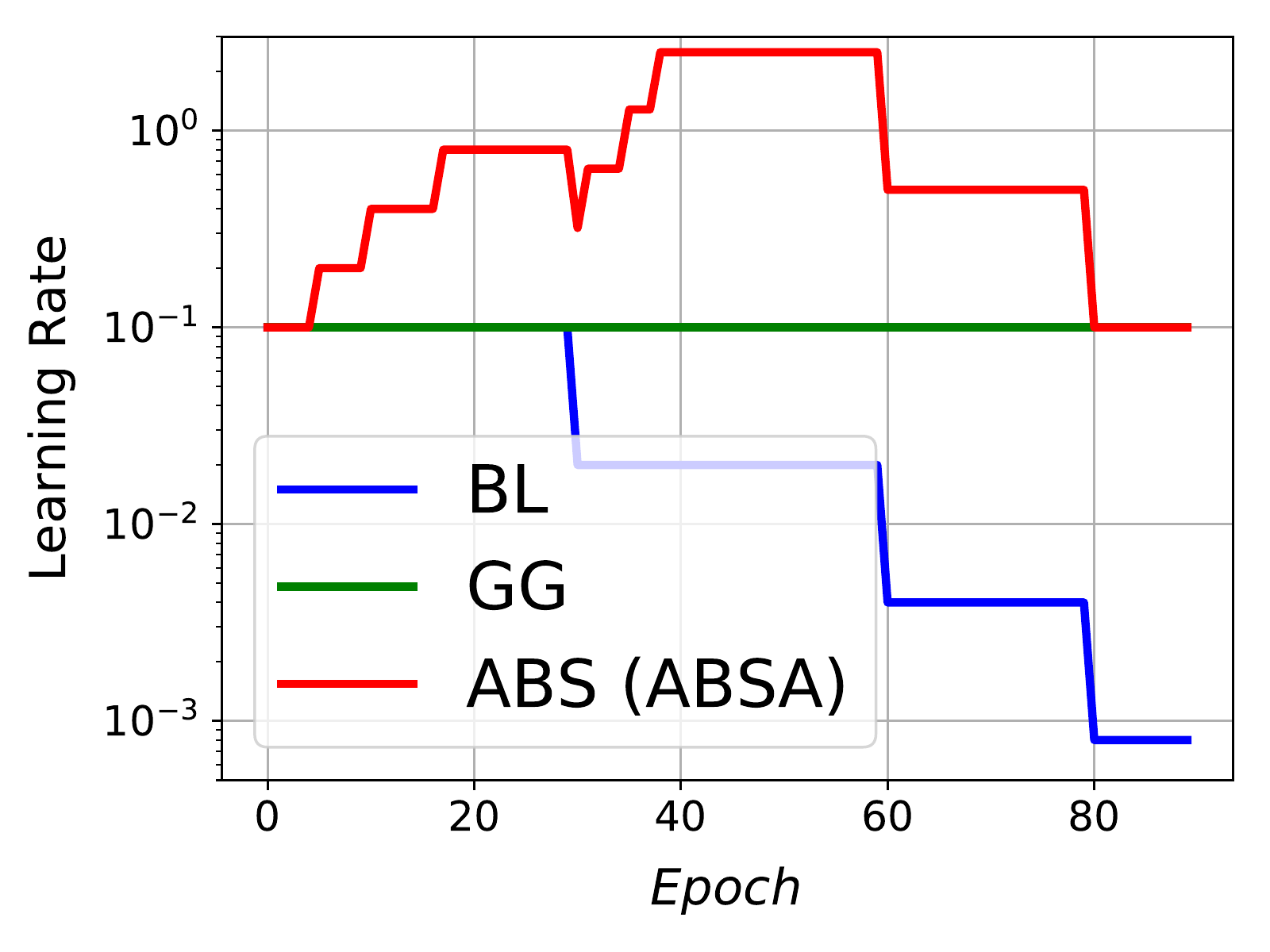}
\includegraphics[width=.45\textwidth]{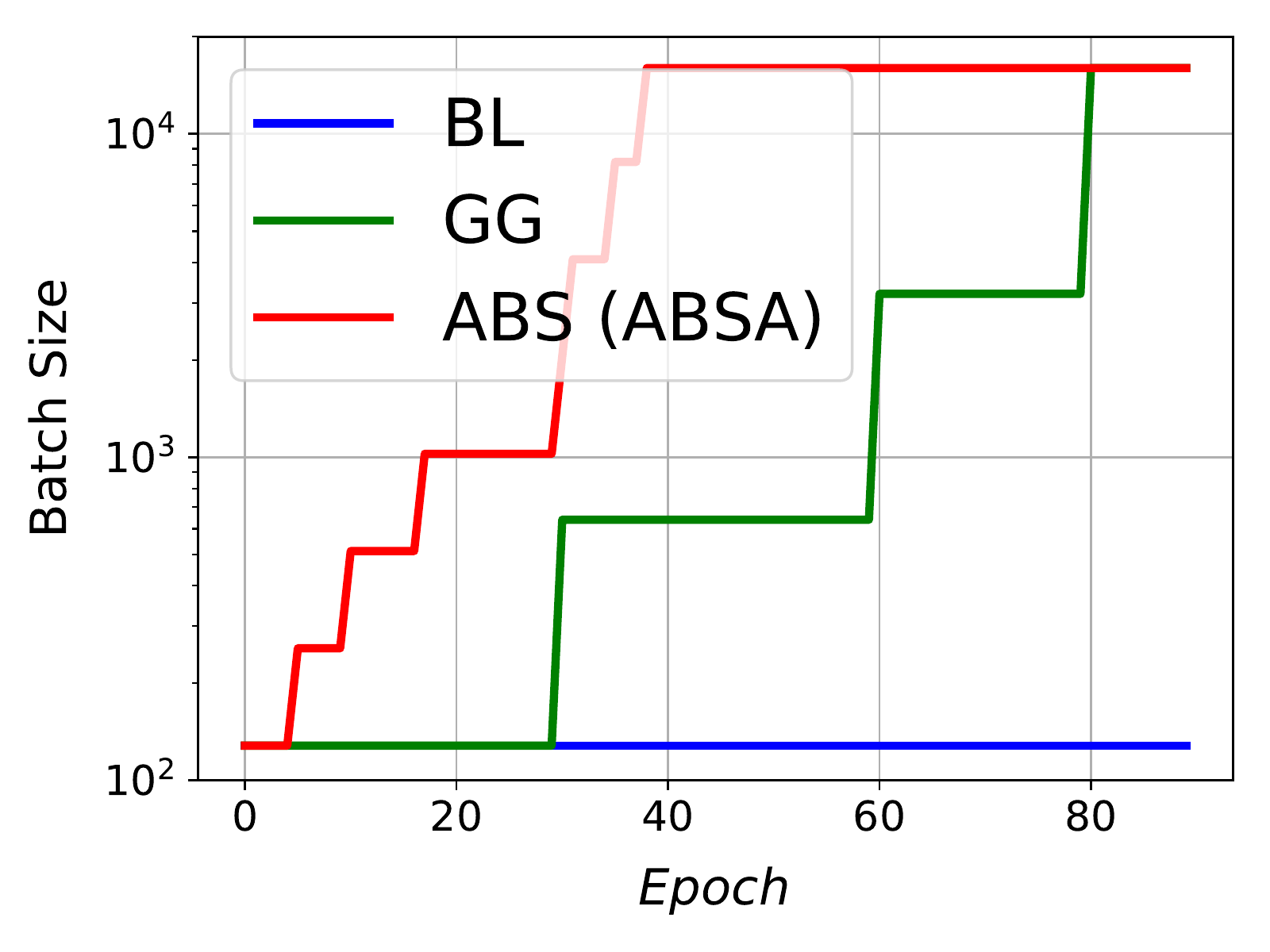}
\end{center}
\caption{{\footnotesize Illustration of learning rate (left) and batch size (right) schedules of adaptive batch size, as a function of training epochs using C2 (Wide-ResNet) model on Cifar-10.
}}
\label{fig:lr_bs_illustration}
\end{figure}
% -------------------

% In the next section, we introduce our ABSA algorithm, and then provide extensive experiments showing
% its efficacy for adaptive batch size training. Moreover, in \S~\ref{sec:time_measurement} we present
% our scalable framework that supports both ABS/ABSA and the method proposed in~\cite{smith2017don} for efficient
% elastic scaling of adaptive workloads.
%%%%%%%%%%%%%%%%%%%%%%%%%%%%%%%%%%%%%%%%%%%%% this is new
\begin{table*}[!htbp]
\caption{\footnotesize 
Accuracy and number of parameter updates of C3 (SqueezeNext) on Cifar-10.}% All methods are trained for 200 epochs. Our method (ABS/ABSA) is in last two columns; and best result of each row is \textbf{Bolded}.}
\label{tab:abs_cifar10_squeeze}
\centering
\small
\begin{tabular}{lcccccccccc} \toprule
                            & \multicolumn{2}{c}{BL}  &  \multicolumn{2}{c}{FB} &   \multicolumn{2}{c}{GG}     & \multicolumn{2}{c}{ABS}  & \multicolumn{2}{c}{ABSA} \\ 
                        \cmidrule{2-3}                  \cmidrule{4-5}              \cmidrule{6-7}    \cmidrule{8-9}                  \cmidrule{10-11}   
    BS                  & {Acc.} & {\# Iters}    & {Acc.} & {\# Iters}          & {Acc.} & {\# Iters}     & {Acc.} & {\# Iters}         & {Acc.} & {\# Iters} \\
    \midrule
\Gc 128                & 92.02 & 78125      & N.A.  & N.A.           & N.A.  & N.A.             & N.A.          & N.A.             & N.A.  & N.A.             \\
\Ga 256                & 91.88 & 39062      & 91.75 & 39062          & 91.84 & 50700            & 91.7          & 40792            &  \textbf{92.11} & 43352    \\
\Gc 512                & 91.68 & 19531      & 91.67 & 19531          & 91.19 & 37050            & \textbf{92.15} & 32428            & 91.61          & 25388   \\
\Ga 1024               & 89.44 &  9766      & 91.23 &  9766          & 91.12 & 31980            & 91.61         & 17046            &  \textbf{91.66} & 23446   \\
\Gc 2048               & 83.17 &  4882      & 90.44 &  4882          & 89.19 & 30030            & 91.57         & 21579            &  \textbf{91.61} & 14027   \\
\Ga 4096               & 73.74 &  2441      & 86.12 &  2441          & 91.83 & 29191            & 91.91         & 18293            &  \textbf{92.07} & 21909   \\
\Gc 8192               & 63.71 &  1220      & 64.91 &  1220          & 91.51 & 28947            & 91.77         & 22802            &  \textbf{91.81} & 16778   \\
\Ga 16384              & 47.84 &   610      & 32.57 &   610          & 90.19 & 28828            & \textbf{92.12} & 17485            & 91.97          & 24361   \\
     \bottomrule 
\end{tabular}
\end{table*}
%%%%%%%%%%%%%%%%%%%%%%%%%%%%%%%%%%%%%%%%%%%%%

\section{Method}
\label{sec:method}

We consider a supervised learning framework where the goal is to minimize a loss function $L(\theta)$:
\small
\begin{equation}
\label{eqn:basic_problem}
L(\theta) = \frac{1}{N} \sum_{i=1}^{N} l(z_i, \theta),
\end{equation}
\normalsize
%\noindent 
where $\theta$ are the model weight parameters,
$Z=(X, Y)$ is the training dataset,
and $l(z, \theta)$ is the loss for a datum $z \in Z$. 
Here, $X$ is the input, $Y$ is the corresponding label, and $N=|Z|$ is the cardinality of the training set.
SGD is typically used to optimize Eq.~(\ref{eqn:basic_problem}) by taking steps of the form:
\small
\begin{equation}
\label{eqn:sgd}
\theta_{t+1} = \theta_t - \eta_t \frac{1}{|B|} \sum_{z \in B} \nabla_\theta l(z, \theta_t),
\end{equation}
\normalsize
%\noindent 
where $B$ is a mini-batch of examples drawn randomly from $Z$, and $\eta_t$ is the step 
size (learning rate) at iteration $t$. 
In the case of large batch size training, the batch size is increased to large values.

As mentioned above, large batch size training often results in convergence to a point with
``sharp'' curvature that often exhibits
poor generalization.
To clearly illustrate this, we show a 3D parametric plot
of the loss landscape when interpolating between the model parameters in the beginning
of training (Epoch 0) and at the end of training (Epoch 90)
with large batch size for the first direction, and interpolate between the same initialization and weights of the model trained with small batch at Epoch 90.
We compute this plot for multiple different models and networks 
(see~\fref{fig:2d_cifar_resnet},~\ref{fig:2d_cifar_resnet_extra_c1} and~\ref{fig:2d_cifar_resnet_extra_c2}).
One can clearly see that the large batch experiments get attracted to a 
sharper landscape point, whereas the small batch training ends in a significantly flatter point.
One popular rationalization of this is that small batch training has sufficient noise to escape such
sharp landscape points.

\subsection{Adaptive Batch Size (ABS) and Adaptive Batch Size Adversarial (ABSA) based on Hessian Information and Adversarial Training}
\label{sec:abs_intro}
 
To address the above problem, \citep{smith2018bayesian} proposed an adaptive batch size scheme which controls SGD noise by increasing batch size instead of adjusting the learning rate.
The adaptive batch size method proposed by~\cite{smith2017don} is shown in~\fref{fig:lr_bs_illustration}.
While this method performs very well, its overall speed up is bottlenecked by the small batch size,
which is used for 30 epochs. (Based on Amdahl's law, even if the rest of the 60 epochs
is trained with infinite speed, the overall training time will be bounded by the first 30 epochs.)
The effect of this bottleneck could be reduced, if we could increase the batch size
sooner. This could be done by manual tuning of the batch size schedule, but as mentioned before
manual tuning is not desired due to additional overhead for the~practitioner.

To address this, we incorporate second-order information to determine when the batch size could be
increased. The corresponding schedule is shown in red in~\fref{fig:lr_bs_illustration}.
The main intuition is to use small batch sizes when the loss landscape has sharper curvature and to use large batch sizes when the loss landscape has flatter
curvature. 
That is, we aim to
keep the SGD noise high to escape the sharp
landscape shown in~\fref{fig:2d_cifar_resnet}. To this end, 
a smaller batch size is used
in regions with a ``sharper'' loss landscape to help avoid
attraction to landscape points with poor generalization.
We then switch to a larger batch size only in regions where the 
loss has a ``flatter'' landscape. As in the hybrid method of~\citep{smith2018bayesian},
we also scale the learning rate proportional to batch size increasing ratio to maintain the noise~level.

However, a key problem here is measuring the loss landscape's curvature.
Naively forming the Hessian operator has a prohibitive computational cost. However,
we can efficiently compute the top eigenvalues of the Hessian operator using power
method, using a matrix free algorithm. 
Let us denote the gradient of $L(\theta)$ w.r.t. $\theta$ by $g$. Then, for a random vector, whose 
dimension is the same as $g$, we~have:
\begin{equation}
    \frac{\partial(g^Tv)}{\partial \theta} = \frac{\partial g^T}{\partial \theta}v + g^T\frac{\partial v}{\partial \theta} = \frac{\partial g^T}{\partial \theta}v = Hv,
\end{equation}
where $H$ is Hessian matrix. Here, the second equation comes from the fact that $v$ 
is independent to $\theta$.
We have developed an efficient implementation to compute curvature information
by using the same automatic differentiation pipeline that is used for
backpropagating the gradient~\cite{hessianflow}. 
The algorithm is shown in~\aref{alg:power_iteration}.
The power iteration algorithm starts with a vector
drawn randomly from a Gaussian distribution, followed by successive application
of the Hessian operator to this vector (so called `matvec'). Thus, instead of computing 
the full Hessian matrix we simply backpropagate this~matvec. 

\begin{algorithm}[t]
\small
\DontPrintSemicolon
\caption{
Power Iteration for Eigenvalue Computation}
\label{alg:power_iteration}
    \SetAlgoLined
    \KwInput{Parameter: $\theta$.
    }
    
    Compute the gradient of $\theta$ by backpropagation, \emph{i.e.}, $g=\frac{d L}{d \theta}$.
    
    Draw a random vector $v$  (same dimension as $\theta$).
    
    Normalize $v$, $v=\frac{v}{\|v\|_2}$
    
    \For(\ \ \quad \quad\quad\quad\quad\tcp*[h]{Power Iteration}){i $=1,2,\ldots, n$}{
        Compute $gv = g^Tv$ \tcp*{Inner product}
        
        Compute $Hv$ by backpropagation, $Hv = \frac{d(gv)}{d\theta}$ \tcp*{Get Hessian vector product}
        
        Normalize and reset $v$, $v = \frac{Hv}{\|Hv\|_2}$
    }
\end{algorithm}

In our approach, the Hessian spectrum is only calculated at the end of every epoch, and thus it has small computational overhead.
We increase the batch size in proportion to how much the Hessian spectrum decays, as compared
to the curvature at initialization\footnote{Note that
sharpness is a relative measure since the absolute value of curvature could be different
for each model/dataset.} 
The pseudo-code for this ABS algorithm is shown in~\aref{alg:abs}.

The second component of our framework is robust optimization.
In~\citep{yao2018hessian}, the authors empirically showed that adversarial training leads to more robust models with 
respect to adversarial perturbation.
An interesting corollary was that, with adversarial 
training, the model converges to regions that are considerably
flatter, as compared to the normal SGD training.
Therefore, we use adversarial training as a regularization term to help regularize against
sharp landscape. Our empirical results have consistently shown improved results.
We refer to this as the Adaptive Batch Size with Adversarial (ABSA) method; see~\aref{alg:abs} for the ABSA pseudo-code. 
In ABSA, we solve a min-max problem instead of a normal empirical risk minimization
problem~\citep{keskar2016large,yao2018hessian}:
\begin{equation}
    \min_\theta \max_{\Delta x \in \mathcal{U}} L(x, y; \theta),
\end{equation}
where $\mathcal{U}$ is an admissibility set for acceptable perturbations (typically restricting the magnitude of the
perturbation). 
Solving this min-max problem for NNs is an intractable 
problem, and thus we approximately solve the maximization 
part through the Fast Gradient Sign Method (FGSM), 
proposed by~\citep{goodfellow6572explaining}. This basically 
corresponds to generating adversarial inputs using one 
gradient ascent step (i.e., the perturbation is computed by 
$\Delta x = \epsilon \sign(\nabla_{x} l(x, y; \theta))$).
Other possible choices are proposed 
by~\citep{thakur2005optimization,carlini2017towards,moosavi2016deepfool,yao2019trust}.%
\footnote{In~\citep{yao2018hessian}, similar behavior was 
observed with other methods for solving the
robust optimization~problem.}

\fref{fig:lr_bs_illustration} illustrates our ABS/ABSA schedule as compared to
a normal training strategy and the adaptive batch size method
of~\citep{smith2017don,devarakonda2017adabatch}. 
As we show in section~\ref{sec:results}, our combined approach (second order and robust optimization) not only achieves better accuracy, but it also requires significantly fewer SGD updates, as compared to~\citep{smith2017don,devarakonda2017adabatch}.

\subsection{Distributed Elastic Training}

Efficient adaptive batch size training is a unique workload that is not well-supported by existing deep learning and cluster computing frameworks. As the batch size increases, more GPUs can be used to parallelize and accelerate the workload. However, all existing solutions for synchronous distributed training assumes a static allocation of GPU resources specified at the beginning of training, while efficient adaptive batch size training forces the resource allocation to be dynamic. Furthermore, though cluster computing offerings such as Kubernetes and AWS offer autoscaling functionality for elastic resource provisioning, there is no existing framework that tightly couples this autoscaling with the SGD training.
To address this, we develop an execution framework for efficient resource-elastic training. We build our framework upon Ray~\cite{moritz2018ray}, a distributed execution engine that provides a distributed training interface and a resource requesting API for increasing the size of the cluster. 
This framework enables efficient GPU allocation from the cloud provider, enabling us to resize training as the target batch size becomes larger. We use PyTorch for the underlying distributed SGD and Hessian based computations. Our distributed Hessian-based computation implements a distributed power iteration, where each process computes Hessian matvec on a subset of input data, and the result is accumulated through an all-reduce collective.

\section{Results}
\label{sec:results}

We evaluate the performance of ABS/ABSA on different datasets (ranging from O(1E5) to O(1E7) training examples)
and multiple NN models.
We compare the baseline SGD performance,
along with other state-of-the-art methods proposed for large batch 
training~\citep{smith2017don,goyal2017accurate}. 
Notice that GG~\citep{smith2017don} and ABS/ABSA have different batch sizes during training. 
Hence the batch size reported in our results represents the maximum batch size during training. 
To allow for a direct comparison we 
also report the number of weight updates in our results (lower is better).
We also report the computed wall-clock time as well using our
Ray based framework for adaptive batch size implementation.

Preferably we would want a higher testing accuracy along with fewer SGD updates. 
Unless otherwise noted, we do \emph{not} change any of the hyper-parameters for
ABS/ABSA.
We use the exact same parameters used in the baseline model, and we do not tailor any parameters
to suit our algorithm. 
A detailed explanation of the different NN models, and the datasets is given in Appendix~\ref{sec:outline_training}.

Section~\ref{sec:results-svhn_and_cifar} shows the result of ABS/ABSA compared to BaseLine (BL), FB~\citep{goyal2017accurate} and GG~\citep{smith2017don}.
Section~\ref{sec:results-imagenet} presents the results on more challenging datasets of TinyImageNet and \textbf{ImageNet}. 
In section~\ref{sec:absa_warm}, we show with slight hyper-parameter tuning on warm-up phase, ABSA can achieve better results. 
Finally, we report the real wall-clock time training by our distribution framework on AWS in section~\ref{sec:time_measurement}.

%%%%%%%%%%%%%%%%%%%%%%%%%%%%%%%%%%%%%%%%%%%%%%%%%%%%%%%%%%%%%%%%%%%%%%%%%%%%%%%%%%%%%%
\begin{figure}[!htbp]
\begin{center}
  \includegraphics[width=.45\textwidth]{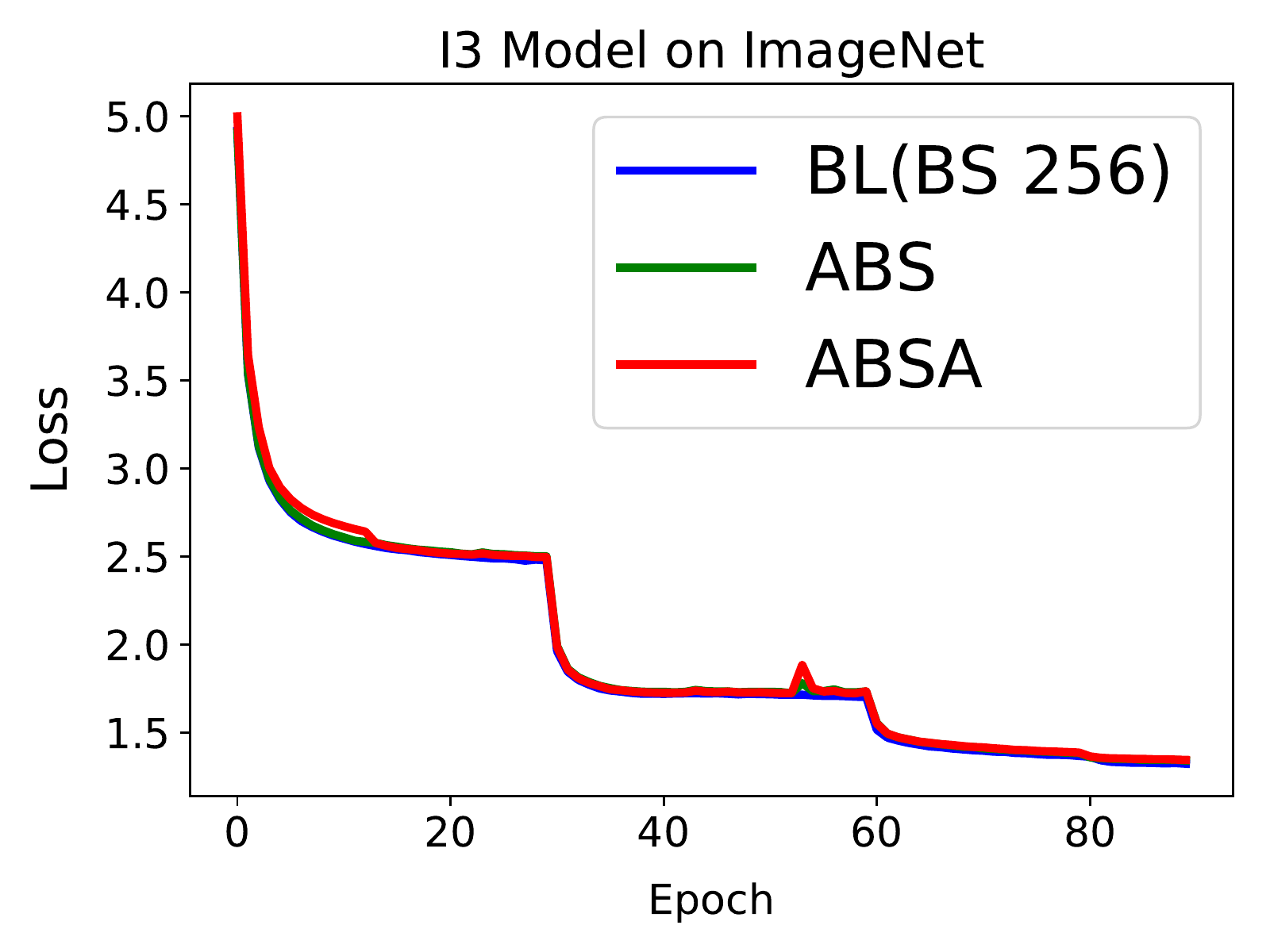}
  \includegraphics[width=.45\textwidth]{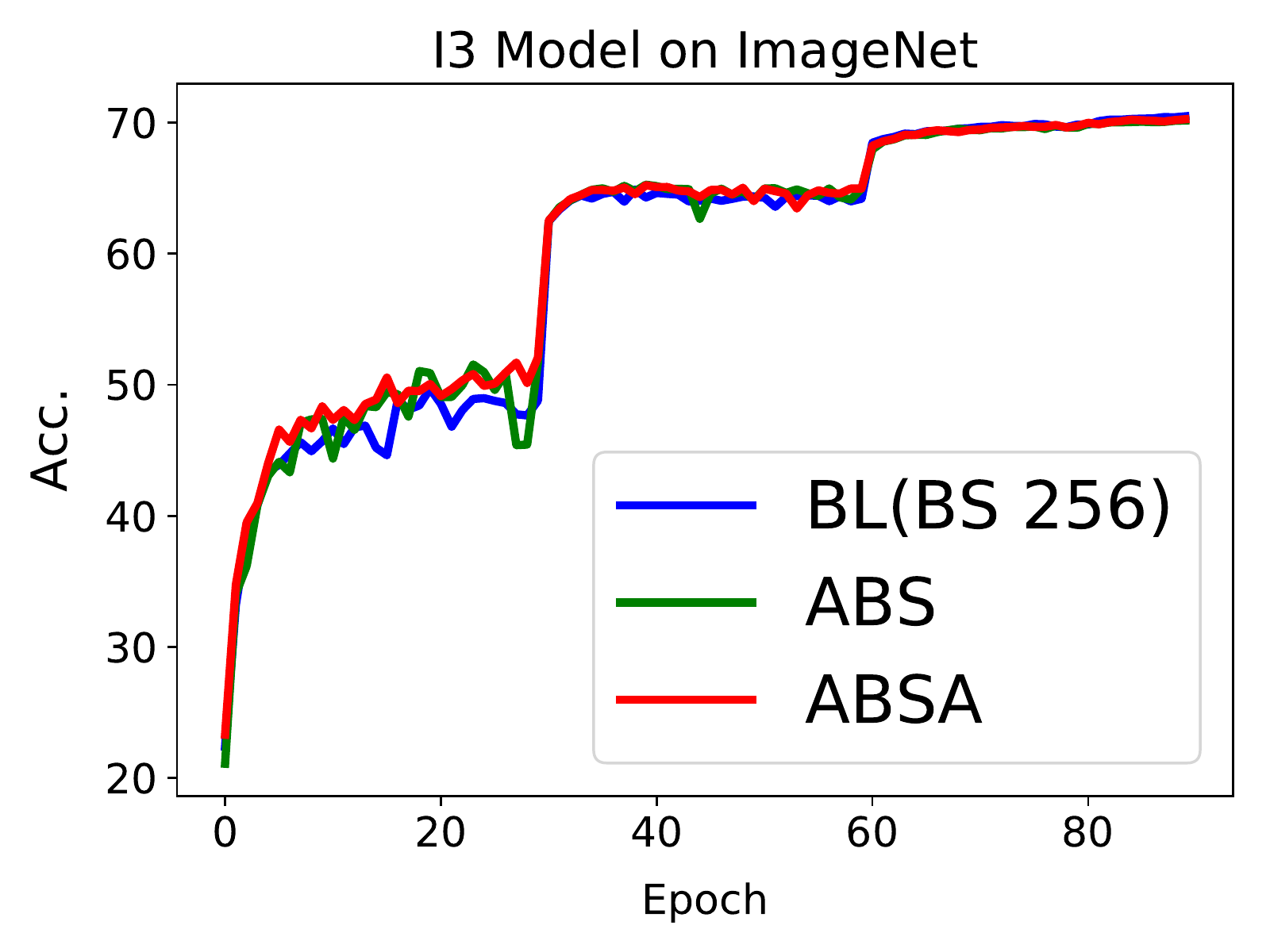}
\end{center}
\caption{
\footnotesize
I3 (ResNet18) model on ImageNet. Training set loss (left) and testing set accuracy (right), evaluated as a function of epochs.
% \red{@Daiyaan we need scaling results for GG, and ABSA for this experiment}
}
\label{fig:imagenet2}
\end{figure}
%%%%%%%%%%%%%%%%%%%%%%%%%%%%%%%%%%%%%%%%%%%%%%%%%%%%%%%%%%%%%%%%%%%%%%%%%%%%%%%%%%%%%%

\subsection{ABS and ABSA for SVHN and Cifar}
\label{sec:results-svhn_and_cifar}

\tref{tab:abs_cifar10_squeeze} and~\tref{tab:abs_svhn}-\ref{tab:abs_cifar100_resnet18} report the test accuracy and the
number of parameter updates for different datasets and models  (see 
Appendix~\ref{sec:extra_result} for~\tref{tab:abs_svhn}-\ref{tab:abs_cifar100_resnet18}). 
First, note the drop in BL accuracy for large batch, confirming the accuracy degradation problem.
Moreover, note that the FB strategy only works well for moderate batch sizes (it diverges for large batch).
However, the GG method has a very consistent 
performance, but its number of parameter updates is usually greater than our method.
Looking at the last two major columns of \tref{tab:abs_cifar10_squeeze} and~\tref{tab:abs_svhn}-\ref{tab:abs_cifar100_resnet18}, the test performance that ABS achieves is similar to BL.
Overall, the number of updates of ABS is 3-10 times smaller than BL with batch size 128. 
Also, note that for most cases ABSA achieves superior accuracy.
This confirms the effectiveness of adversarial training combined with the second order information.
Furthermore, we show a similar 3D parametric plot for ABSA algorithm 
in~\fref{fig:2d_cifar_resnet},~\ref{fig:2d_cifar_resnet_extra_c1} and~\ref{fig:2d_cifar_resnet_extra_c2}.
Note that ABSA avoids
the sharp landscape of large batch and is able to converge to a landscape with similar
curvature as that of BL with small batch training.

Furthermore, we show the loss landscape of model obtained throughout the training process along the dominant eigenvalue~\cite{yao2018hessian} in~\fref{fig:loss_landscape_1d_resnet18} in Appendix~\ref{sec:extra_result}.
It can be clearly seen that throughout training larger batches tend to get stuck/attracted
to areas with larger curvature, while small batch SGD and ABSA are able to avoid them.

% \vspace{-2mm}
\subsection{ABSA for TinyImageNet and ImageNet}
\label{sec:results-imagenet}

Here, we report the ABSA method on more challenging datasets, specifically, TinyImageNet and ImageNet.
We use the exact same hyper-parameters in our algorithm as for Cifar/SVHN, even though tuning them could potentially be preferable for us.
All TinyImageNet results are in the Appendix~\ref{sec:extra_result}; see~\fref{fig:tinyimagenet} and \tref{tab:abs_tinyimagenet}.

Due to the limited computational resources, we only test ABSA and BL (baseline with small batch) for ImageNet, and we report results in~\fref{fig:imagenet} (see Appendix~\ref{sec:extra_result}).
We first start with I2 model (AlexNet) whose baseline
requires $450K$ parameter updates (i.e., SGD iterations), reaching $56.32\%$ validation accuracy. 
For ABSA, the final validation accuracy is $56.40\%$, with only $76K$ parameter updates. 
The maximum batch size reached by ABSA is $16,384$, with initial batch size $256$. 

We also test I3 (ResNet18) model on ImageNet, as shown in~\fref{fig:imagenet2}.
The BL with batch of 256 reaches $70.46\%$ validation accuracy with $450K$ parameter updates.
The final validation accuracies of ABS and ABSA are $70.15\%$ and $70.24\%$, respectively, both with $66,393$ parameter updates.
The maximum batch size reached by ABS and ABSA is $16,384$ with initial batch size $256$.
If GG schedule is implemented, the total number of parameter updates would have been $166K$.
(Due to the limitation of resource, we do not run GG.)

\subsection{ABSA with Warm-up Tuning}\label{sec:absa_warm}
To \textit{effectively} speed up training, large batch should not require
extensive hyper-parameter tuning.
However, a recent trend (mainly motivated by industry labs) is focused on showing
how fast training on a benchmark task could be finished on a supercomputer
~\cite{you2017scaling,mikami2018imagenet,ying2018image,goyal2017accurate, smith2018bayesian}.
This is mostly
not possible with academic resources. However, to illustrate how ABSA's performance
could be boosted by additional hyper-parameter tuning, we solely focus on the warmup schedule (and keep
all the other hyper-parameters such as $\alpha,~\beta$ the same).
For the warmup tuning, our goal is to ramp up to large batches quickly.
On Cifar-10 with C1 model, we increase the initial batch size to $1920$ (as compared to baseline batch size $128$), and we gradually increase the learning rate in five epochs, as in~\cite{goyal2017accurate}. The final  accuracy of this mixed schedule is 83.24\% with 784 SGD updates (compared to  83.04\% for baseline with $35K$ SGD updates, which is $48.1\times$ smaller).

On ImageNet with I3 model, we increase the initial batch size to $4096$ (as compared to 
baseline batch size $256$) with largest batch size $16K$, and we gradually increase
the learning rate in 20 epochs, and train for a total of 100 epochs.
The final accuracy is $70.04\%$ with $14.8K$ SGD updates (as compared to 70.4\% for baseline with 450k SGD updates, which is $28.8\times$ smaller).
 
We emphasize that these results are achieved with moderate hyper-parameter tuning though only 10 trials; and also we did not tune any other hyper-parameters introduced by ABSA. It is quite feasible that with an industrial infrastructure and tuning of learning rate, momentum, and ABSA's parameters, the results could significantly be improved. 
 
\subsection{Training Time Measurement}\label{sec:time_measurement}
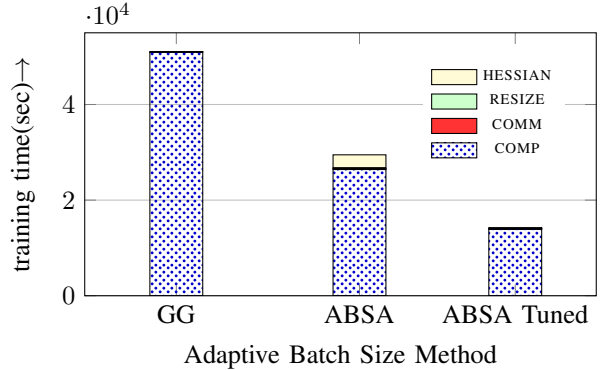
\begin{figure}[ht] %% <<< Weak scaling sphere
  \begin{minipage}[b]{0.9\linewidth}\centering
    % \vspace{5pt}
    % \hspace{-45pt}
    \begin{tikzpicture} %<<<
      \begin{semilogxaxis}
        [ ybar stacked,
          legend style={draw=none, at={(0.8,0.9)}, anchor=north,legend columns=1,reverse legend, /tikz/every even column/.append style={column sep=0.3cm}},
          xtick=data,xticklabels={GG,ABSA,ABSA Tuned}, xlabel=Adaptive Batch Size Method,
          ytick={}, ymajorgrids,ylabel=training time(sec)$\rightarrow$,
        y label style={at={(0.06,0.5)}},
          xmin=0.5, xmax=24,
          ymin=0, ymax=55000,
          bar width=20pt,
          width=3.3in,
          height=2.0in]

        \addplot [area legend, draw=black,fill=blue!20, pattern = crosshatch dots,pattern color=blue!80!black
%        nodes near coords, nodes near coords align={vertical},
        ] coordinates {
        (   1,50965.0197)
        (   4,26404)
        (   13, 13935)};
        \addplot [area legend, draw=black,fill= red!80] coordinates {
        (   1,53.7389)
        (   4, 230)
        (   13, 58)};
        
        \addplot [area legend, draw=black,fill= green!20,] coordinates {
        (   1,39.41)
        (   4, 94.967)
        (   13, 39.43)};
        
        \addplot [area legend, draw=black,fill= yellow!20] coordinates {
        (   1,0)
        (   4, 2746.225)
        (   13, 220.048)};

        \makeatletter
        \pgfplots@stacked@isfirstplottrue
        \makeatother
        \addplot [forget plot,draw=none] coordinates{
        (   1,0)
        (   4,0)
        (   13, 0)};

        \legend{{\scriptsize \sc{comp}}, {\scriptsize \sc{comm}}, {\scriptsize \sc{resize}}, {\scriptsize \sc{hessian}}}
      \end{semilogxaxis}
    \end{tikzpicture} %>>>
  \end{minipage}
  \hfill
  %\caption{Weak scalability on Stampede up to 2k compute nodes for flow through 250 spheres.}
  \caption{\footnotesize
  Breakdown of the different methods' runtimes for our Ray based adaptive batch size framework. In particular, note the very small overhead for resizing the cluster, as well as the small Hessian computation overhead for ABSA.}
  \label{fig:scalingbar}
\end{figure}

To show the scaling of our framework, we test both GG and ABSA for ResNet18 ImageNet training.
We set the batch size per GPU to be 256 and run our experiments on AWS EC2 using p3.16xlarge instances. 
We report SGD computation/communication, cost of resizing the cluster, and Hessian 
computation time, as shown in~\fref{fig:scalingbar} and~\tref{tab:scalingtab}.
We see that resizing and communication time 
only introduce slight overhead to computation time. 
GG only reduces the training time from 125K to 51K seconds because it is bottlenecked  by the first phase with 256 batch size, as shown in~\fref{fig:lr_bs_illustration}. 
ABSA further reduces the training time to 29.4K seconds.
Note that the Hessian spectrum computation only accounts for
9.3\% of the entire training time.
The main bottleneck of ABSA Hessian
computation is that at the very beginning, we only allow 1 or 2 GPUs to compute Hessian
information, which has to be done with gradient accumulation. 
For ABSA Tuned, where we tune the warm-up phase, the time can be reduced to 14.2K. Again,
note the small overhead of the Hessian computation, as compared to the total training time.
 
%%%%%%%%%%%%%%%%%%%%%%%%%%%%%%%%%%%%%%%%%%%%%%%%%%%%%%%%%%%%%%%%%%%%%%%%%%%%%%%%%%%%%%
 %% >>>
%%%%%%%%%%%%%%%%%%%%%%%%%%%%%%%%%%%%%%%%%%%%%%%%%%%%%%%%%%%%%%%%%%%%%%%%%%%%%%%%%%%%%%
\begin{table}[!h]
\centering
\small 
\setlength\tabcolsep{1.pt}
\begin{tabular}{lcccccccccccccc} \toprule
Method              & Comp    & Comm  & Resize           & Hess  & Total   & Speedup \\
\midrule
\ha Baseline         &   125073 &  N/A &  N/A & N/A    & 125073      &   1x    \\
\midrule
\hc  GG            & 50965    & 54     & 40    & N/A   & 51059     &    2.45x \\ % & 1.00$\times$\\                       
\midrule
\ha ABSA            & 26404    & 230    & 95    & 2746  & 29475   &       4.24x \\ % & 8.00$\times$
\hc ABSA Tuned      & 13935    & 58     & 39    & 220   & \bf{14252}    &    \bf{8.78x} \\
\bottomrule 
\end{tabular}
\caption{
\footnotesize 
Breakdown of ResNet18 training time on ImageNet
for GG~\cite{smith2018bayesian}, ABSA,
and ABSA Tuned. For the latter, we only tuned the warm-up phase and did not
tune any other hyper-parameters. We report
SGD computation time (Comp), SGD communication time (Comm),
cost of resizing the cluster (Resize), overhead of Hessian computation and
communication (Hess), and the total training time (Total) to reach baseline accuracy.}
\label{tab:scalingtab}
\end{table}
%%%%%%%%%%%%%%%%%%%%%%%%%%%%%%%%%%%%%%%%%%%%%%%%%%%%%%%%%%%%%%%%%%%%%%%%%%%%%%%%%%%%%%

\section{Conclusion}

In this work, we address two major challenges for adaptive batch size training of NNs.
First, we developed a new framework based on Ray that allows efficient dynamic scaling of the cluster with negligible resize cost. 
Even so, the speed up of existing adaptive batch size methods is limited by the initial portion of the training with small batch size. 
Thus, we presented a Hessian based method which allows more rapid increase of the batch size during the initial phase.
This reduces the impact of this phase on overall speed up.
We extensively test these and other related methods on five datasets with multiple NN models (AlexNet, ResNet, Wide ResNet and SqueezeNext).
To do so, we performed two sets of experiments, one without any hyper-parameter tuning, and
one where we only tuned the warm-up phase of the training.
In both cases, our ABSA method enables one to increase batch size more rapidly resulting in fewer number of SGD iterations.
Furthermore, we showed scaling results of our proposed method and measured run times on AWS for both pior work~\cite{smith2017don} as well as our ABSA method.
To enable reproducibility, we open source our distribution tools in~\cite{hessianflow}.

\section*{Acknowledgments}
This would was supported by a gracious fund from Intel corporation. We would like
to thank Intel VLAB team for providing us with access to their computing cluster.
We also gratefully acknowledge the support of NVIDIA Corporation with the donation of the Titan Xp GPU used for this research.
Furthermore, MWM would like to acknowledge ARO, DARPA, NSF, and ONR for providing partial support of this work.
% \clearpage
{\small
\bibliographystyle{plain}
\bibliography{ref}
}
% \clearpage
\onecolumn
\appendix

% \vspace{-2mm}
\section{Convergence Rate of ABS}
Here, we provide a straightforward proof that ABS algorithm does converge for strongly convex problems. This proof is a basic modification of SGD's convergence proof with
constant batch size.
Based on an assumption about the loss (Assumption~\ref{ass:1} in Appendix~\ref{app:thm}), it is not hard to prove the following theorem.
\begin{mytheorem}\label{thm}
Under Assumption~\ref{ass:1}, assume at step $t$, the batch size used for parameter update is $b_t$, the step size is $b_t\eta_0$, where $\eta_0$ is fixed and satisfies,
\small
\begin{equation}
0<\eta_0 \leq \frac{1}{L_g(M_v+B_{\max})},
\end{equation}
\normalsize
where $B_{\max}$ is the maximum batch size during training.
Then, with $\theta_0$ as the initialization, the expected optimality gap satisfies the following inequality,
\small
\begin{equation}\label{eqn:theory_result}
\E[L(\theta_{t+1})] - L_*\leq \prod_{k=1}^t (1-b_k\eta_0c_s)(L(\theta_0)-L_*-\frac{\eta_0L_gM}{2c_s}) + \frac{\eta_0L_gM}{2c_s}.
\end{equation}
\normalsize
\end{mytheorem}

From Theorem~\ref{thm}, if $b_t \equiv 1$, the convergence rate for $t$ steps, based on~\eqref{eqn:theory_result}, is $(1-\eta_0c_s)$. 
However, the convergence rate of~\aref{alg:abs} becomes $\prod_{k=1}^t (1-b_k\eta_0c_s)$, 
where $1\leq b_k\leq B_{max}$. With an adaptive $b_t$~\aref{alg:abs} can converge faster than basic SGD.
We show empirical results for a logistic regression problem in the Appendix~\ref{app:thm}, which is a simple convex problem.

\subsection{Proof of Theorem}\label{app:thm}
For a finite sum objective function $L(\theta)$, i.e.,~\eqref{eqn:basic_problem}, we assume that:
\begin{assumption}\label{ass:1} 
The objective function $L(\theta)$ satisfies:
\begin{itemize}[noitemsep,topsep=0pt,parsep=0pt,partopsep=0pt,leftmargin=*]
\item $L(\theta)$ is continuously differentiable and the gradient function of $L$ is Lipschitz continuous with Lipschitz constant $L_g$, i.e.
\begin{equation}
\|\nabla L(\theta_1)-\nabla L(\theta_2)\| \leq L_g\|\theta_1-\theta_2\|,~~~~~~~~~~\text{for all $\theta_1$ and $\theta_2$}.
\end{equation}

\item $L(\theta)$ is strongly convex, i.e.,  there exists a constant $c_s>0$ s.t.
\begin{equation}
L(\theta_2) \geq L(\theta_1) + \nabla L(\theta_1)^T(\theta_2-\theta_1) + \frac12 c_s\|\theta_1-\theta_2\|^2,~~~~~~~~~~\text{for all $\theta_1$ and $\theta_2$}.
\end{equation}
Also, the global minima of $L(\theta)$ is achieved at $\theta_*$ and $L(\theta_*)=L_*$.

\item Each gradient of each individual $l_i(z_i)$ is an unbiased estimation of the true gradient, i.e.
\begin{equation}
\E[\nabla l_i(z_i, \theta)] = \nabla L(\theta),~~~~~~~~~~\text{for all $i$}.
\end{equation}

\item There exist scalars $M\geq0$ and $M_v\geq 0$ s.t.
\begin{equation}
\V(\nabla l_i(z_i, \theta)) \leq M + M_v\|\nabla L(\theta)\|,~~~~~~~~~~\text{for all $i$},
\end{equation}
where $\V(\cdot)$ is the variance operator, i.e.
\[
    \V(\nabla l_i(z_i, \theta)) = \E[\|\nabla l_i(z_i, \theta)\|^2] - \|\E[\nabla l_i(z_i, \theta)]\|^2.
\]
\end{itemize}
\end{assumption}

From the Assumption~\ref{ass:1}, it is not hard to get,
\begin{equation}
\E[\| \nabla l_i(z_i, \theta)\|^2] \leq M + M_g\|\nabla L(\theta)\|^2,
\end{equation}
with $M_g=M_v+1$.

With Assumption~\ref{ass:1}, the following two lemmas could be found in any optimization reference, e.g. \cite{bottou2018optimization}. 
We give the proofs here for completeness.

\begin{lemma}\label{lemma:2} Under Assumption~\ref{ass:1}, after one iteration of stochastic gradient update with step size $\eta_t$ at $\theta_t$, we have
\begin{equation}
\E[L(\theta_{t+1})] - L(\theta_t) \leq -(1-\frac12 \eta_tL_gM_g) \eta_t \|\nabla L(\theta_t)\|^2 + \frac12 \eta_t^2L_gM,
\end{equation}
where $\theta_{t+1}=\theta_t-\eta_t\nabla l_i(\theta, z_i)$ for some $i$.
\end{lemma}

%\paragraph{Proof.}
\begin{proof}
With the $L_g$ smooth of $L(\theta)$, we have
\begin{align*}
\E[L(\theta_{t+1})] - L(\theta_t) &\leq -\eta_t \nabla L(\theta_t)\E[\nabla l_i(\theta, z_i)] + \frac12\eta_t^2L_g\E[\|\nabla l_i(\theta, z_i)\|^2]\\
                                  &\leq -\eta_t \|\nabla L(\theta_t)\|^2 + \frac12\eta_t^2L_g(M+M_g\|\nabla L(\theta_t)\|^2).
\end{align*}
From above, the result follows.
\end{proof}

\begin{lemma}\label{lemma:3} Under Assumption~\ref{ass:1}, for any $\theta$, we have
\begin{equation}
2c_s(L(\theta)-L_*) \leq \|\nabla L(\theta)\|^2.
\end{equation}
\end{lemma}
\begin{proof}
Let 
\[
    h(\bar\theta) = L(\theta) + \nabla L(\theta)^T (\bar\theta-\theta) + \frac12c_s\|\bar\theta-\theta\|^2.
\]
Then $h(\bar\theta)$ has a unique global minima at $\bar\theta_* = \theta-\frac{1}{c_s}\nabla L(\theta)$ with $h(\bar\theta_*)=L(\theta)-\frac{1}{2c_s}\|\nabla L(\theta)\|^2$. Using the strong convexity of $L(\theta)$, it follows
\[
    L(\theta_*) \geq L(\theta) + \nabla L(\theta)^T(\theta_*-\theta) + \frac12 c_s\|\theta-\theta_*\|_2^2 =h(\bar\theta_*)=L(\theta)-\frac{1}{2c_s}\|\nabla L(\theta)\|^2.
\]

\end{proof}

The following lemma is trivial, we omit the proof here. 
\begin{lemma}\label{lemma:4} Let $L_B(\theta) = \frac{1}{|B|} \sum_{z_i \in B}l_i(\theta, z_i)$. Then the variance of $\nabla L_B(\theta)$ is bounded by
\begin{equation}
\V(\nabla L_B(\theta)) \leq M/|B| + M_v\|\nabla L(\theta)\|/|B|,~~~~~~~~~~\text{for all B}.
\end{equation}
\end{lemma}

\paragraph{Proof of Theorem~\ref{thm}}
Given these lemmas, we now proceed with the proof of Theorem~\ref{thm}.

\begin{proof}
Assume the batch used at step t is $b_t$, according to Lemma~\ref{lemma:2} and~\ref{lemma:4}, 
\begin{align*}
\E[L(\theta_{t+1})] - L(\theta_t) 
&\leq -(1-\frac12 b_t\eta_0L_g(\frac{M_v}{b_t}+1)) b_t\eta_0 \|\nabla L(\theta_t)\|^2 + \frac12 (b_t\eta_0)^2L_g\frac{M}{b_t}\\
& \leq  -(1-\frac12\eta_0L_g(M_v+b_t)) b_t\eta_0 \|\nabla L(\theta_t)\|^2 + \frac12 b_t\eta_0^2L_gM\\
& \leq  -(1-\frac12\eta_0L_g(M_v+B_{\max})) b_t\eta_0 \|\nabla L(\theta_t)\|^2 + \frac12 b_t\eta_0^2L_gM\\
&\leq -\frac12 b_t\eta_0 \|\nabla L(\theta_t)\|^2 + \frac12 b_t\eta_0^2L_gM \\
&\leq -b_t\eta_0c_s (L(\theta_t)-L_*) + \frac12 b_t\eta_0^2L_gM,
\end{align*}
where the last inequality is from Lemma~\ref{lemma:3}. This yields
\begin{align*}
\E[L(\theta_{t+1})] - L_* 
&\leq L(\theta_t) -b_t\eta_0c_s (L(\theta_t)-L_*) + \frac12 b_t\eta_0^2L_gM - L_* \\
& = (1-b_t\eta_0c_s)(L(\theta_t)-L_*) + \frac12 b_t\eta_0^2L_gM. \\
\end{align*}
It is not hard to see,
\[
    \E[L(\theta_{t+1})] - L_* -\frac{\eta_0L_gM}{2c_s} \leq  (1-b_t\eta_0c_s)(L(\theta_t)-L_*-\frac{\eta_0L_gM}{2c_s}),
\]
from which it follows:
\[
\E[L(\theta_{t+1})] - L_* -\frac{\eta_0L_gM}{2c_s} \leq \prod_{k=1}^t (1-b_k\eta_0c_s)(L(\theta_0)-L_*-\frac{\eta_0L_gM}{2c_s}).
\]
Therefore, 
\[
    \E[L(\theta_{t+1})] - L_*\leq \prod_{k=1}^t (1-b_k\eta_0c_s)(L(\theta_0)-L_*-\frac{\eta_0L_gM}{2c_s}) + \frac{\eta_0L_gM}{2c_s}.
\]
\end{proof}

\begin{figure}[tbp]
\begin{center}
  \includegraphics[width=.45\textwidth]{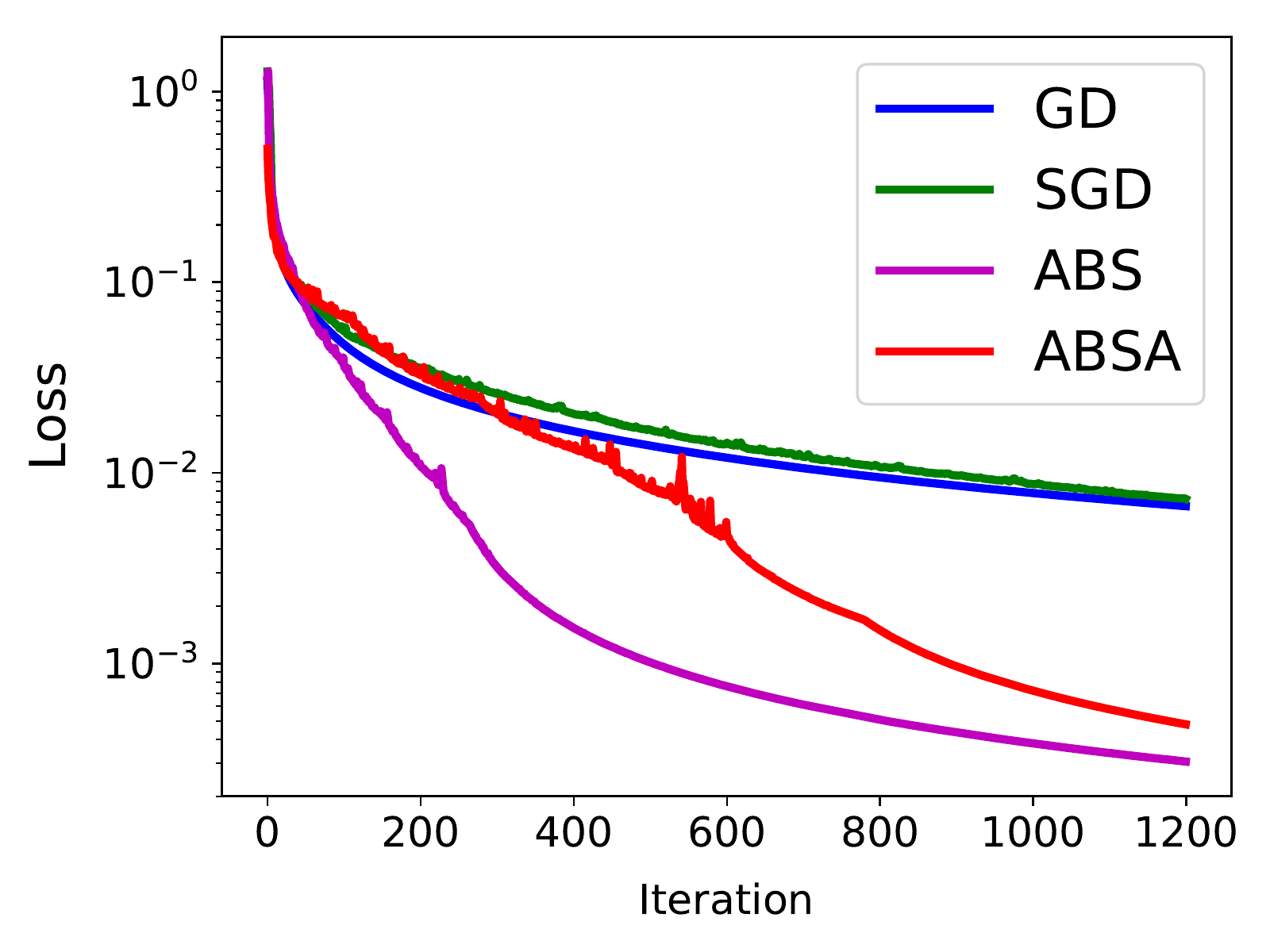}
  \includegraphics[width=.45\textwidth]{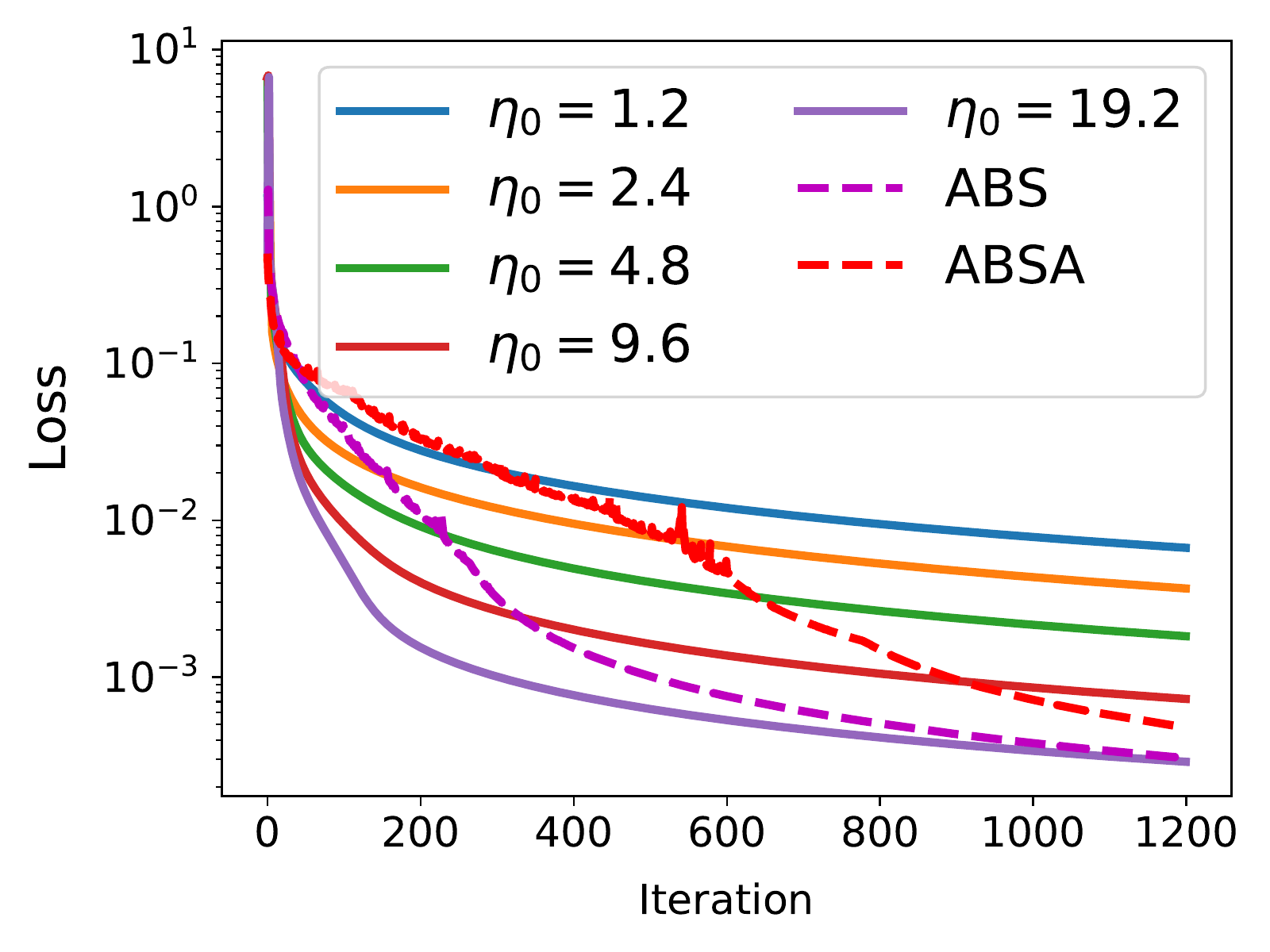}
\end{center}
\caption{Logistic regression model result. The left figure shows the training loss as a function of iterations for full gradient, SGD, ABS and ABSA. The right figure shows the result of ABS/ABSA compared to full gradient with different learning rate.}
\label{fig:mushroom}
\end{figure}

We show a toy example of binary logistic regression on mushroom classification dataset\footnote{https://www.kaggle.com/uciml/mushroom-classification}. We split the whole dataset to 6905 for training and 1819 for validation. $\eta_0=1.2$ for SGD with batch size 100 and full gradient descent. We set $100\leq b_t\leq 3200$ for our algorithm, i.e. ABS. Here we mainly focus on the training losses of different optimization algorithms. The results are shown in Figure~\ref{fig:mushroom}. In order to see if $\eta_0$ is not an optimal step size of full gradient descent, we vary $\eta_0$ for full gradient descent; see results in Figure~\ref{fig:mushroom}. 

%%%%%%%%%%%%%%%%%%%%%%%%%%%%%%%%%%%%%%%%%%%%%%%%%%%%%%%%%%%%%%%%%%%%%%%%%%%%%%%%%%%%%%
%%%%%%%%%%%%%%%%%%%%%%%%%%%%%%%%%%%%%%%%%%%%%%%%%%%%%%%%%%%%%%%%%%%%%%%%%%%%%%%%%%%%%%

\section{Training Details}\label{sec:outline_training}
In this section, we give the detailed outline of our training datasets, models, strategy as well as hyper-parameter used in Alg~\ref{alg:abs}.

\textbf{Dataset.}
We consider the following datasets.

\begin{itemize}[noitemsep,topsep=0pt,parsep=0pt,partopsep=0pt,leftmargin=*]

\item 
\textbf{SVHN.} 
The original SVHN~\citep{netzer2011reading} dataset is small. 
However, in this paper, we choose the additional dataset, which contains more than 500k samples, as our training dataset. 

\item 
\textbf{Cifar.} 
The two Cifar (i.e., Cifar-10 and Cifar-100) datasets~\citep{krizhevsky2009learning} have same number of images but different number of classes.

\item 
\textbf{TinyImageNet.} 
TinyImageNet consists of a subset of ImangeNet images~\citep{deng2009imagenet}, which contains 200 classes. 
Each of the class has 500 training and 50 validation images.\footnote{In some papers, this validation set is sometimes referred to as a test set.}
The size of each image is $64\times 64$. 

\item 
\textbf{ImageNet.} 
The ILSVRC 2012 classification dataset~\citep{deng2009imagenet} consists of 1000 images classes, with a total of 1.2 million training images and 50,000 validation images. 
During training, we crop the image to $224 \times 224$.
\end{itemize}

\textbf{Model Architecture.}
We implement the following convolution NNs. 
When we use data augmentation, it is exactly same the standard data augmentation scheme as in the corresponding model.

\begin{itemize}[noitemsep,topsep=0pt,parsep=0pt,partopsep=0pt,leftmargin=*]
\item \textbf{S1.} AlexNet like model on SVHN as same as~\citep[C1]{yao2018hessian}. We train it for 20 epochs with initial learning rate $0.01$, and decay a factor of 5 at epoch 5, 10 and 15. There is no data augmentation. The batch size to compute Hessian information is $128$.
\item \textbf{C1.} ResNet20 on Cifar-10 dataset~\citep{he2016deep}. We train it for 90 epochs with initial learning rate $0.1$, and decay a factor of 5 at epoch 30, 60, 80. There is no data augmentation. The batch size to compute Hessian information is $128$.
\item \textbf{C2.} Wide-ResNet 16-4 on Cifar-10 dataset~\citep{zagoruyko2016wide}. We train it for 90 epochs with initial learning rate $0.1$, and decay a factor of 5 at epoch 30, 60, 80. There is no data augmentation. The batch size to compute Hessian information is $128$.
\item \textbf{C3.} SqueezeNext on Cifar-10 dataset~\citep{gholami2018squeezenext}. We train it for 200 epochs with initial learning rate $0.1$, and decay a factor of 5 at epoch 60, 120, 160. Data augmentation is implemented. The batch size to compute Hessian information is $128$.
\item \textbf{C4.} 2.0-SqueezeNext (twice width) on Cifar-10 dataset~\citep{gholami2018squeezenext}. We train it for 200 epochs with initial learning rate $0.1$, and decay a factor of 5 at epoch 60, 120, 160. Data augmentation is implemented.
\item \textbf{C5.} ResNet18 on Cifar-100 dataset~\citep{he2016deep}. We training it for 160 epochs with initial learning rate $0.1$, and decay a factor of 10 at epoch 80, 120. Data augmentation is implemented. The batch size to compute Hessian information is $1024$.
\item \textbf{I1.} ResNet50 on TinyImageNet dataset~\citep{he2016deep}. We training it for 120 epochs with initial learning rate $0.1$, and decay a factor of 10 at epoch 60, 90. Data augmentation is implemented. The batch size to compute Hessian information is $2048$.
\item \textbf{I2.} AlexNet on ImageNet dataset~\citep{krizhevsky2012imagenet}. We train it for 90 epochs with initial learning rate $0.01$, and decay it to $0.0001$ quadratically at epoch 60, then keeps it as $0.0001$ for the remaining 30 epochs. Data augmentation is implemented. The batch size to compute Hessian information is $4096$.
\item \textbf{I3.} ResNet18 on ImageNet dataset~\citep{he2016deep}. We train it for 90 epochs with initial learning rate $0.1$, and decay a factor of 10 at epoch 30, 60 and 80. Data augmentation is implemented. The batch size to compute Hessian information is $4096$.
\end{itemize}

\subsection{Training Strategy:}
\label{sec:training_strategy}
We use the following training strategies
\begin{itemize}[noitemsep,topsep=0pt,parsep=0pt,partopsep=0pt,leftmargin=*]
\item \textbf{BL.} Use the standard training procedure. 
\item \textbf{FB.} Use linear scaling rule \citep{goyal2017accurate} with warm-up stage.
\item \textbf{GG.} Use increasing batch size instead of decay learning rate \citep{smith2017don}. 
\item \textbf{ABS.} Use our adaptive batch size strategy \emph{without} adversarial training.
\item \textbf{ABSA.} Use our adaptive batch size strategy \emph{with} adversarial training.
\end{itemize}

For adversarial training, the adversarial data are generated using Fast Gradient Sign Method (FGSM)~\citep{goodfellow6572explaining}.
The hyper-parameters in Alg.~\ref{alg:abs} ($\alpha$ and $\beta$) are chosen to be $2$, $\kappa=10$, $\epsilon_{adv}=0.005$, $\gamma=20\%$, and $\omega=2$ for all the experiments.
The only change is that for SVHN, the frequency to compute Hessian information is $65536$ training examples as compared to one epoch, due to the small number of total training epochs (only 20).

\section{Approximate Hessian}
\label{sec:block_hessian}

\begin{figure}[!htbp] %[tbp]
\begin{center}
  \includegraphics[width=.45\textwidth]{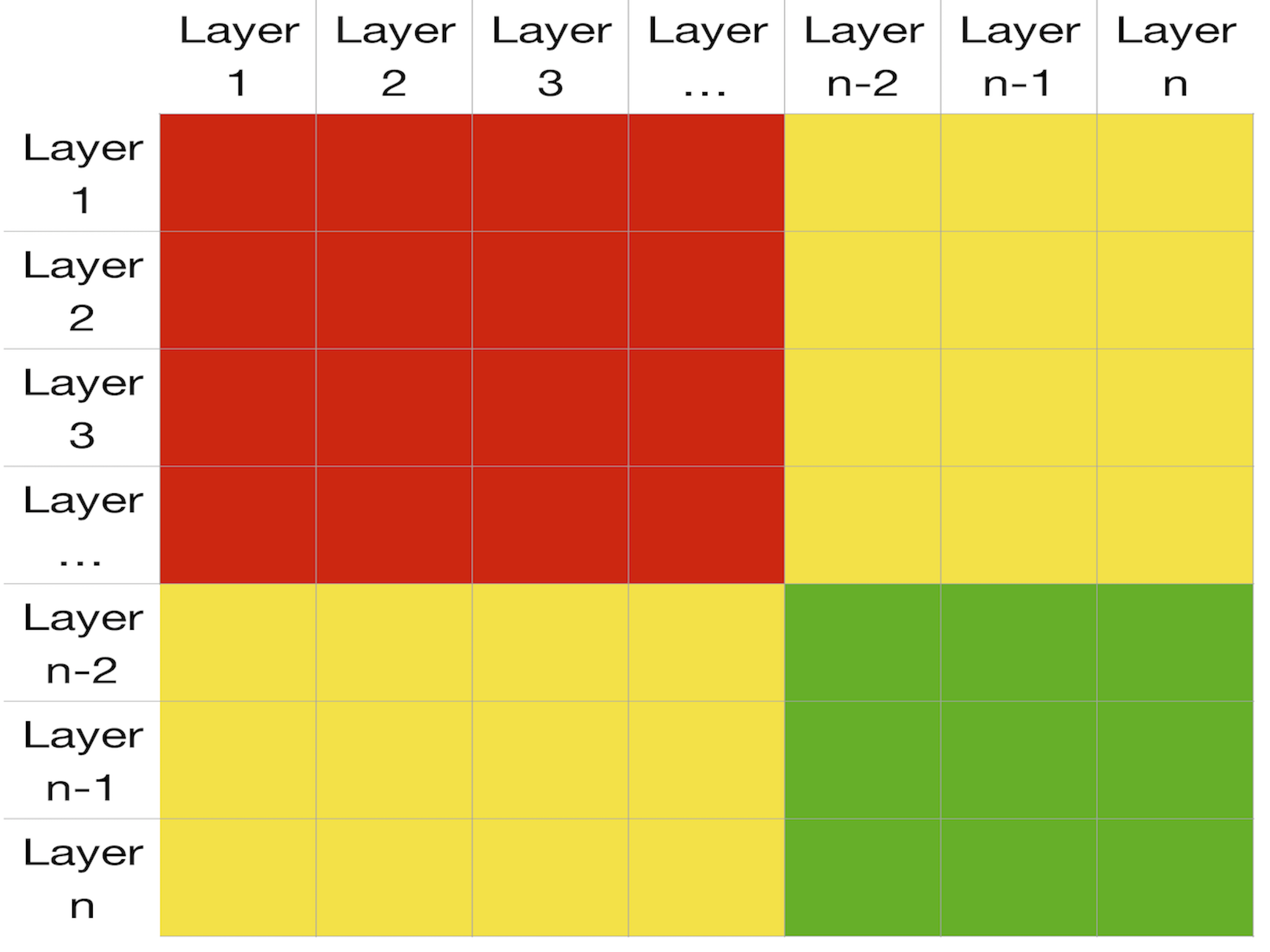}
  \includegraphics[width=.45\textwidth]{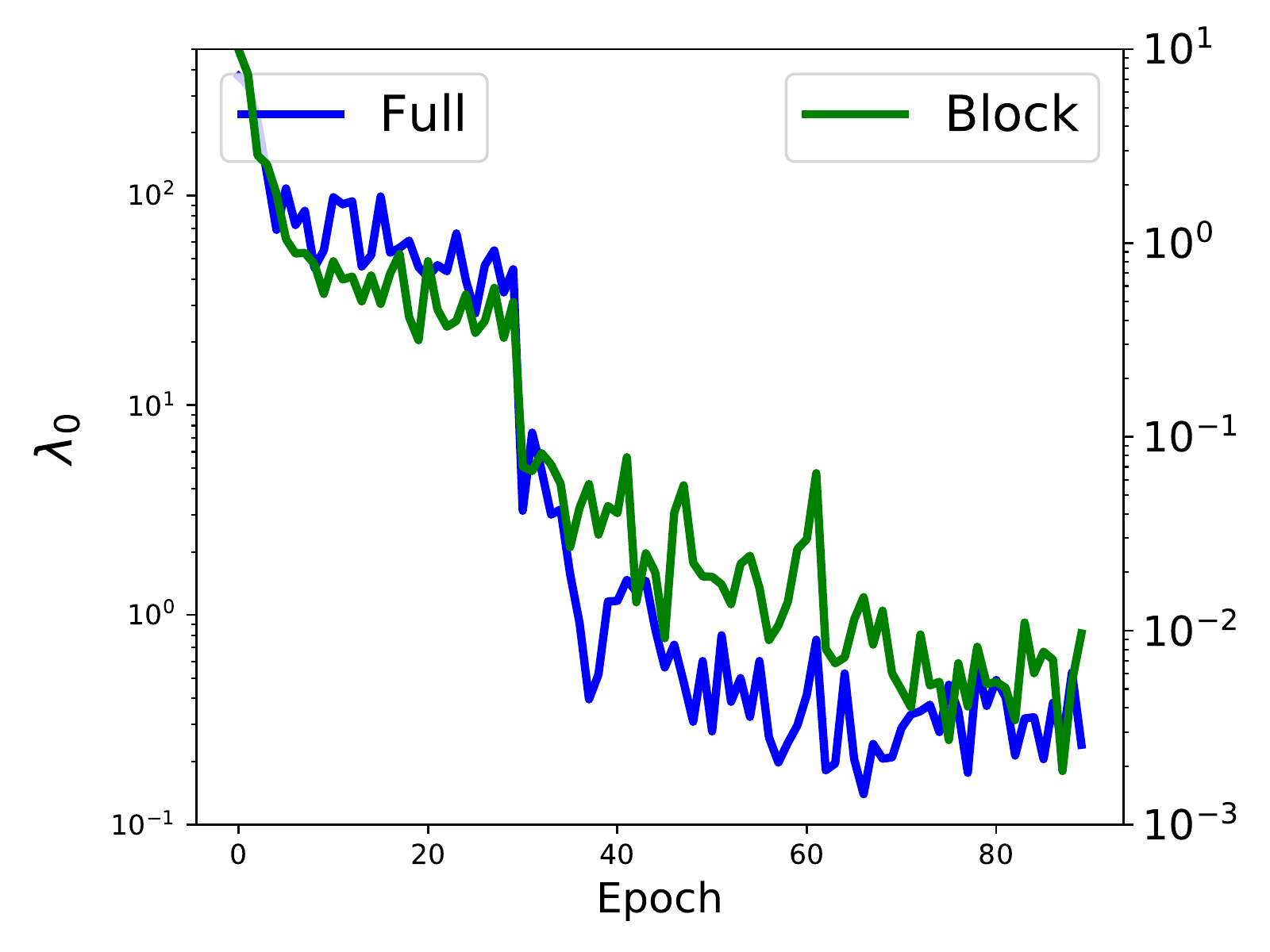}
\end{center}
\caption{Illustration of block Hessian (left). Instead computing the top eigenvalue of whole Hessian, we just compute the 
eigenvalue of the green block. 
Top eigenvalues of Block of C1 (right) on Cifar-10. The block Hessian is computed by the last two layers of C1. 
The maximum batch size of C1 is 16000.
The full Hessian is based on BL with batch size of 128.  }
\label{fig:block_hessian}
\end{figure}

One of the limitations of second order methods is the additional computational cost for computing the top Hessian eigenvalue.
If we use the full Hessian operator, the second backpropagation needs to be done all the way to the first layer of NN. 
For deep networks this could lead to high computational cost. 
Here, we empirically explore whether we could use approximate second order information, and in particular we test a block Hessian approximation as illustrated in~\fref{fig:block_hessian}.
The block approximation corresponds to only analyzing the Hessian of the last few layers.

In~\fref{fig:block_hessian}, we plot the trace of top eigenvalues of full Hessian and block Hessian for C1 model. 
Although the top eigenvalue of block Hessian has more variance than that of full Hessian, the overall trends are similar for C1. 
The test performance of C1 on Cifar-10 with block Hessian is $84.82\%$ with 4600 parameter updates (as compared to $84.42\%$ for full Hessian ABSA).
The test performance of C4 on Cifar-100 with block Hessian is $68.01\%$ with 12500 parameter updates (as compared to $68.43\%$ for full Hessian ABSA).
These results suggest that using a block Hessian to estimate the trend of the full Hessian might be a good choice to overcome computation cost, but a more detailed analysis is needed.
Other approaches such as sketching might be another possible direction to
reduce this overhead~\cite{gupta2019oversketched}.

%%%%%%%%%%%%%%%%%%%%%%%%%%%%%%%%%%%%%%%%%%%%%%%%%%%%%%%%%%%%%%%%%%%%%%%%%%%%%%%%%%%%%%

\section{Additional empirical results}\label{sec:extra_result}
In this section, we present additional empirical results that
were discussed in Section 4 (i.e., Table~\ref{tab:abs_cifar10_squeeze}-\ref{tab:abs_tinyimagenet}, and~\fref{fig:imagenet2}).

\begin{table}[!htbp]
\caption{Accuracy and the number of parameter updates of S1 on SVHN.}
\label{tab:abs_svhn}
\centering
\begin{tabular}{lcccccccccc} \toprule
                      & \multicolumn{2}{c}{BL}  &  \multicolumn{2}{c}{FB} &   \multicolumn{2}{c}{GG}     & \multicolumn{2}{c}{ABS}  & \multicolumn{2}{c}{ABSA} \\ 
                        \cmidrule{2-3}                  \cmidrule{4-5}              \cmidrule{6-7}    \cmidrule{8-9}                  \cmidrule{10-11}   
    BS                  & {Acc.} & {\# Iters}    & {Acc.} & {\# Iters}          & {Acc.} & {\# Iters}     & {Acc.} & {\# Iters}         & {Acc.} & {\# Iters} \\
    \midrule
\Gc    128             & 94.90 & 81986         & N.A.  & N.A.                  & N.A.           & N.A.            & N.A.   & N.A.               & N.A.            & N.A.     \\                  
\Ga    512             & 94.76 & 20747         & 95.24 & 20747                 & 95.49          & 51862           & 95.65  & 25353              & \textbf{95.72}  & 24329    \\              
\Gc   2048             & 95.17 & 5186          & 95.00 & 5186                  & 95.59          & 45935           & 95.51  & 10562              & \textbf{95.82}  & 16578    \\              
\Ga   8192             & 93.73 & 1296          & 19.58 & 1296                  & \textbf{95.70} & 44407           & 95.56  & 14400              & 95.61           & 7776     \\             
\Gc  32768             & 91.03 & 324           & 10.00  & 324                   & 95.60          & 42867           & 95.60  & 7996               & \textbf{95.90}  & 12616    \\              
\Ga 131072             & 84.75 & 81            & 10.00  & 81                    & 95.58          & 42158           & 95.61  & 11927              & \textbf{95.92}  & 11267    \\             
     \bottomrule 
\end{tabular}
\end{table}

\begin{table}[!htbp]
\caption{Accuracy and the number of parameter updates of C1 on Cifar-10.}
\label{tab:abs_cifar10_resnet18}
\centering
\begin{tabular}{lcccccccccc} \toprule
                      & \multicolumn{2}{c}{BL}  &  \multicolumn{2}{c}{FB} &   \multicolumn{2}{c}{GG}     & \multicolumn{2}{c}{ABS}  & \multicolumn{2}{c}{ABSA} \\ 
                        \cmidrule{2-3}                  \cmidrule{4-5}              \cmidrule{6-7}    \cmidrule{8-9}                  \cmidrule{10-11}   
    BS                  & {Acc.} & {\# Iters}    & {Acc.} & {\# Iters}          & {Acc.} & {\# Iters}     & {Acc.} & {\# Iters}         & {Acc.} & {\# Iters} \\
    \midrule
\Gc 128                & 83.05 & 35156                &  N.A.           &  N.A.             & N.A.    & N.A.              & N.A.    & N.A.              & N.A.            & N.A. \\
\Ga 640                & 81.01 & 7031                 & \textbf{84.59}  &  7031             & 83.99   & 16380             & 83.30   & 10578             & 84.52           & 9631 \\
\Gc 3200               & 74.54 & 1406                 &  78.70          &  1406             & 84.27   & 14508             & 83.33   & 6375              & \textbf{84.42}  & 5168 \\
\Ga 5120               & 70.64 & 878                  &  74.65          &  878              & 83.47   & 14449             & 83.83   & 6575              & \textbf{85.01}  & 6265 \\
\Gc 10240              & 68.75 & 439                  &  30.99          &  439              & 83.68   & 14400             & 83.56   & 5709              & \textbf{84.29}  & 7491 \\
\Ga 16000              & 67.88 & 281                  &  10.00          &  281              & 84.00   & 14383             & 83.50   & 5739              & \textbf{84.24}  & 5357 \\
     \bottomrule 
\end{tabular}
\end{table}

\begin{table}[!htbp]
\caption{Accuracy and the number of parameter updates of C2 on Cifar-10.}
\label{tab:abs_cifar10_wresnet}
\centering
\begin{tabular}{lcccccccccc} \toprule
                      & \multicolumn{2}{c}{BL}  &  \multicolumn{2}{c}{FB} &   \multicolumn{2}{c}{GG}     & \multicolumn{2}{c}{ABS}  & \multicolumn{2}{c}{ABSA} \\ 
                        \cmidrule{2-3}                  \cmidrule{4-5}              \cmidrule{6-7}    \cmidrule{8-9}                  \cmidrule{10-11}   
    BS                  & {Acc.} & {\# Iters}    & {Acc.} & {\# Iters}          & {Acc.} & {\# Iters}     & {Acc.} & {\# Iters}         & {Acc.} & {\# Iters} \\
    \midrule
\Gc 128                & 87.64 & 35156                &  N.A.   &  N.A.             & N.A.    & N.A.              & N.A.    & N.A.              & N.A.           & N.A. \\
\Ga 640                & 86.20 & 7031                 &  87.9   &  7031             & 87.84   & 16380             & 87.86   & 10399             & \textbf{89.05} & 10245 \\
\Gc 3200               & 82.59 & 1406                 &  73.2   &  1406             & 87.59   & 14508             & 88.02   & 5869              & \textbf{89.04} & 4525 \\
\Ga 5120               & 81.40 & 878                  &  63.27  &  878              & 87.85   & 14449             & 87.92   & 7479              & \textbf{88.64} & 5863 \\
\Gc 10240              & 79.85 & 439                  &  10.00  &  439              & 87.52   & 14400             & 87.84   & 5619              & \textbf{89.03} & 3929 \\
\Ga 16000              & 81.06 & 281                  &  10.00  &  281              & 88.28   & 14383             & 87.58   & 9321              & \textbf{89.19} & 4610 \\
     \bottomrule 
\end{tabular}
\end{table}

\begin{table}[!htbp]
\caption{Accuracy and the number of parameter updates of C4 on Cifar-10.}
\label{tab:abs_cifar10_squeeze2_0}
\centering
\begin{tabular}{lcccccccccc} \toprule
                      & \multicolumn{2}{c}{BL}  &  \multicolumn{2}{c}{FB} &   \multicolumn{2}{c}{GG}     & \multicolumn{2}{c}{ABS}  & \multicolumn{2}{c}{ABSA} \\ 
                        \cmidrule{2-3}                  \cmidrule{4-5}              \cmidrule{6-7}    \cmidrule{8-9}                  \cmidrule{10-11}   
    BS                  & {Acc.} & {\# Iters}    & {Acc.} & {\# Iters}          & {Acc.} & {\# Iters}     & {Acc.} & {\# Iters}         & {Acc.} & {\# Iters} \\
    \midrule
\Gc 128                 & 93.25   & 78125        & N.A.    & N.A.               & N.A.       & N.A.       & N.A.     & N.A.             & N.A.       & N.A. \\
\Ga 256                 & 93.81   & 39062        & 89.51   & 39062              & 93.53      & 50700      & 93.47    & 43864            & \bf{93.76}      & 45912\\
\Gc 512                 & 93.61   & 19531        & 89.14   & 19531              & 93.87      & 37058      & 93.54    & 25132            & \bf{93.57}      & 28494\\
\Ga 1024                & 91.51   & 9766         & 88.69   & 9766               & 93.21      & 31980      &\bf{93.40}& 19030            &    {93.26}      & 22741\\
\Gc 2048                & 87.90   & 4882         & 88.03   & 4882               & 93.54      & 30030      & 93.40    & 21387            & \bf{93.50}      & 23755\\
\Ga 4096                & 80.77   & 2441         & 86.21   & 2441               & 93.32      & 29191      & 93.55    & 14245            & \bf{93.82}      & 20557\\
% \Gc 8192                & 71.74   & 1220         & 55.72   & 1220               & 93.14      & 28947      &     &                       &       &\\
% \Ga 16384               & 55.32   & 610          & 16.12   & 610                & 93.60      & 28828      &     &                       &       &\\
%% this last two rows of our results are not good. It is similar to the model on cifar-100. 
     \bottomrule 
\end{tabular}
\end{table}

\begin{table}[!htbp]
\caption{Accuracy and the number of parameter updates of C5 on Cifar-100. }
\label{tab:abs_cifar100_resnet18}
\centering
\begin{tabular}{lcccccccccc} \toprule
                      & \multicolumn{2}{c}{BL}  &  \multicolumn{2}{c}{FB} &   \multicolumn{2}{c}{GG}     & \multicolumn{2}{c}{ABS}  & \multicolumn{2}{c}{ABSA} \\ 
                        \cmidrule{2-3}                  \cmidrule{4-5}              \cmidrule{6-7}    \cmidrule{8-9}                  \cmidrule{10-11}   
    BS                  & {Acc.} & {\# Iters}    & {Acc.} & {\# Iters}          & {Acc.} & {\# Iters}     & {Acc.} & {\# Iters}         & {Acc.} & {\# Iters} \\
    \midrule
\Gc 128              & 67.67  & 62500          & N.A.            & N.A.             & N.A.             & N.A.    & N.A.  & N.A.             & N.A.            & N.A    \\
\Ga 256              & 67.12  & 31250          & \textbf{67.89}  & 31250            & 66.79            & 46800   & 67.71 & 33504            & 67.32           & 33760  \\
\Gc 512              & 66.47  & 15625          & 67.83           & 15625            & 67.74            & 39000   & 67.68 & 32240            & \textbf{67.87}  & 24688  \\
\Ga1024              & 64.7   & 7812           & 67.72           & 7812             & 67.17            & 35100   & 65.31 & 22712            & \textbf{68.03}  & 13688  \\
\Gc2048              & 62.91  & 3906           & 67.93           & 3906             & 67.76            & 33735   & 64.69 & 25180            & \textbf{68.43}  & 12103  \\
     \bottomrule 
\end{tabular}
\end{table}

% %%%%%%%%%%%%%%%%%%%%%%%%%%%%%%%%%%%%%%%%%%%%%%%%%%%%%%%%%%%%%%%%%%%%%%%%%%%%%%%%%%%%%%
\begin{table*}[!htbp]
\caption{Accuracy and the number of parameter updates of I1 on TinyImageNet.}% All methods are trained for 120 epochs. Our method (ABS/ABSA) is in last two columns; and best result of each row is \textbf{Bolded}.}
\label{tab:abs_tinyimagenet}
\centering
\begin{tabular}{lcccccccc} \toprule
                      & \multicolumn{2}{c}{BL}  &  \multicolumn{2}{c}{FB} &   \multicolumn{2}{c}{GG}      & \multicolumn{2}{c}{ABSA} \\ 
                        \cmidrule{2-3}                  \cmidrule{4-5}              \cmidrule{6-7}    \cmidrule{8-9}                   
    BS                  & {Acc.} & {\# Iters}    & {Acc.} & {\# Iters}          & {Acc.} & {\# Iters}     & {Acc.} & {\# Iters}        \\
    \midrule
\Gc 128              & 60.41     & 93750              &  N.A.      & N.A.      & N.A.   & N.A.                                & N.A.   & N.A. \\
\Ga 256              & 58.24     & 46875              &  59.82     & 46875     & 60.31  & 70290                               & \textbf{61.28}  & 60684 \\
\Gc 512              & 57.48     & 23437              &  59.28     & 23437     & 59.94  & 58575                               & \textbf{60.55}  & 51078 \\
\Ga1024              & 54.14     & 11718              &  59.62     & 11718     & 59.72  & 52717                               & \textbf{60.72}  & 19011 \\
\Gc2048              & 50.89     &  5859              &  59.18     &  5859     & 59.82  & 50667                               & \textbf{60.43}  & 17313  \\
\Ga4096              & 40.97     &  2929              &  58.26     &  2929     & 60.09  & 49935                               & \textbf{61.14}  & 22704 \\
\Gc 8192             & 25.01     &  1464              &  16.48     &  1464     & 60.00  & 49569                               & \textbf{60.71}  & 22334 \\
\Ga16384             & 10.21     &   732              &  0.50       &   732     & 60.37  & 48995                               & \textbf{60.71}  & 20348 \\
     \bottomrule 
\end{tabular}
\end{table*}

% %%%%%%%%%%%%%%%%%%%%%%%%%%%%%%%%%%%%%%%%%%%%%%%%%%%%%%%%%%%%%%%%%%%%%%%%%%%%%%%%%%%%%%

%%%%%%%%%%%%%%%%%%%%%%%%%%%%%%%%%%%%%%%%%%%%%%%%%%%%%%%%%%%%%%%%%%%%%%%%%%%%%%%%%%%%%%
\begin{figure}[!htbp] %[tbp]
\begin{center}
  \includegraphics[width=.45\textwidth]{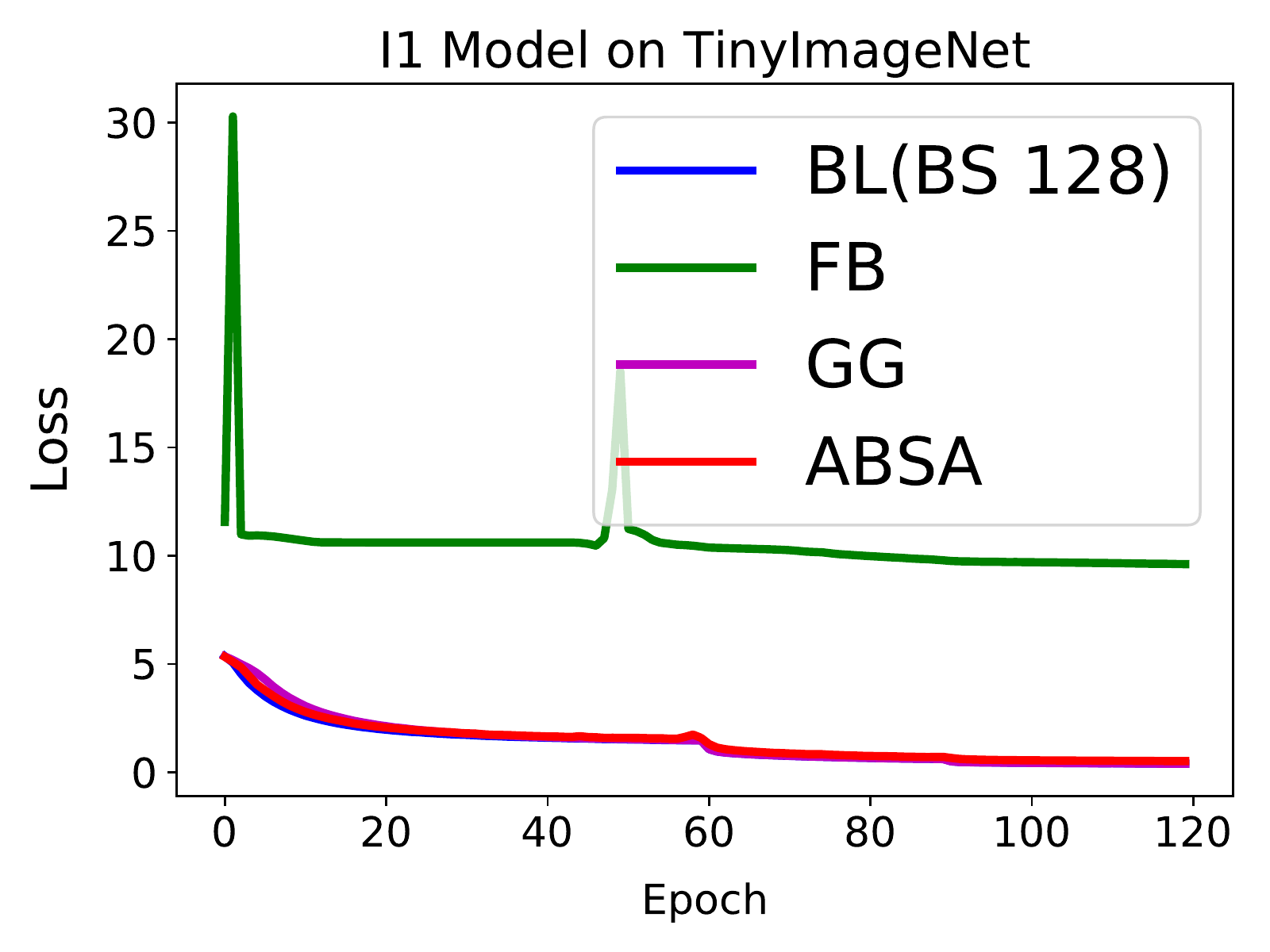}
  \includegraphics[width=.45\textwidth]{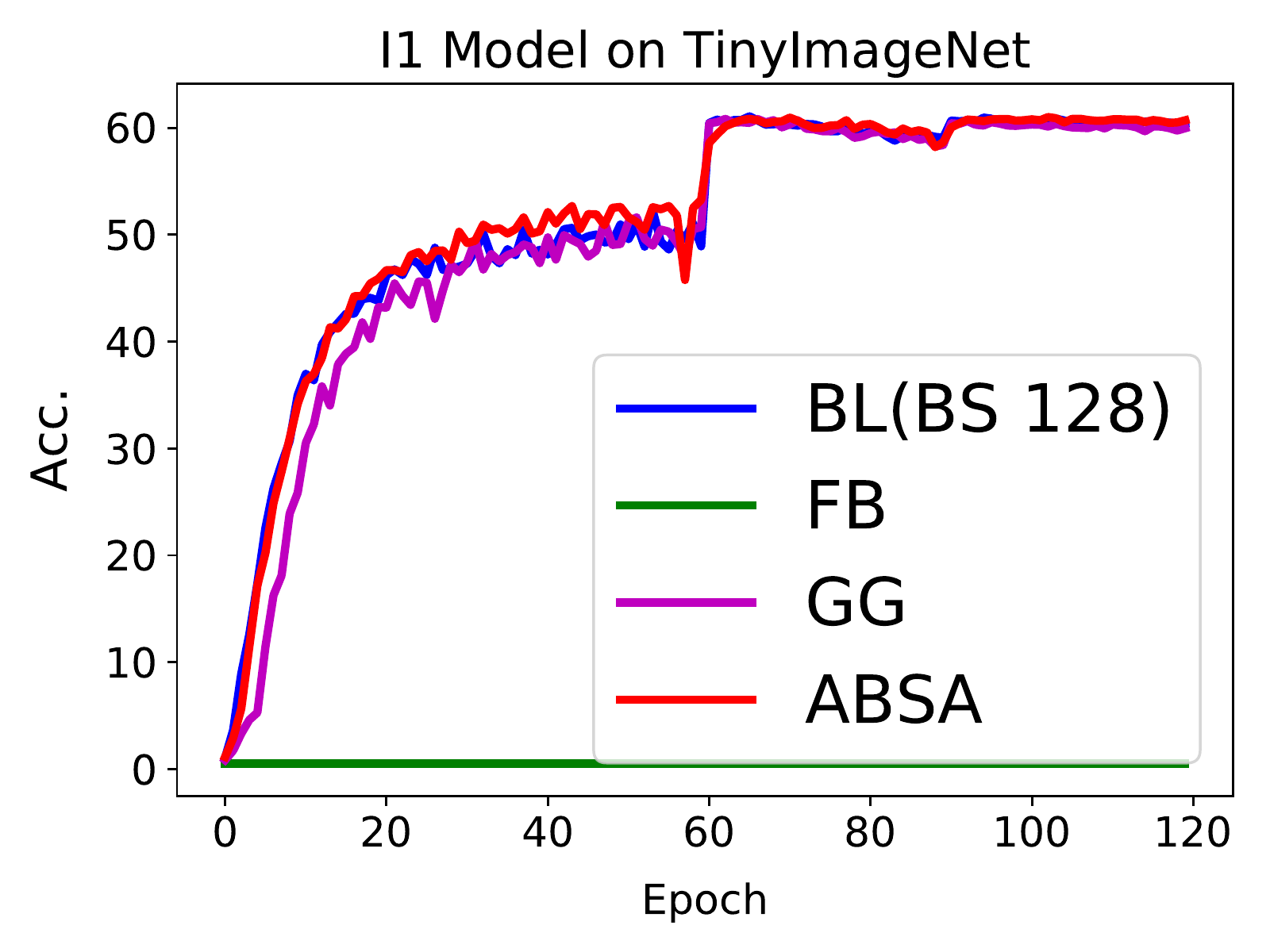}
\end{center}
\caption{I1 model on TinyImageNet. Training set loss (left), and testing set accuracy (right), evaluated as a function of epochs. 
All results correspond to batch size 16384 (please see~\tref{tab:abs_tinyimagenet} for details).
As one can see,
from epoch 60 to 80, the test performance drops due to overfitting.
However, ABSA achieves the best performance with apparently less overfitting (it has higher training loss). 
}
\label{fig:tinyimagenet}
\end{figure}
%%%%%%%%%%%%%%%%%%%%%%%%%%%%%%%%%%%%%%%%%%%%%%%%%%%%%%%%%%%%%%%%%%%%%%%%%%%%%%%%%%%%%%

%%%%%%%%%%%%%%%%%%%%%%%%%%%%%%%%%%%%%%%%%%%%%%%%%%%%%%%%%%%%%%%%%%%%%%%%%%%%%%%%%%%%%%
\begin{figure}[!htbp] %[tbp]
\begin{center}
  \includegraphics[width=.45\textwidth]{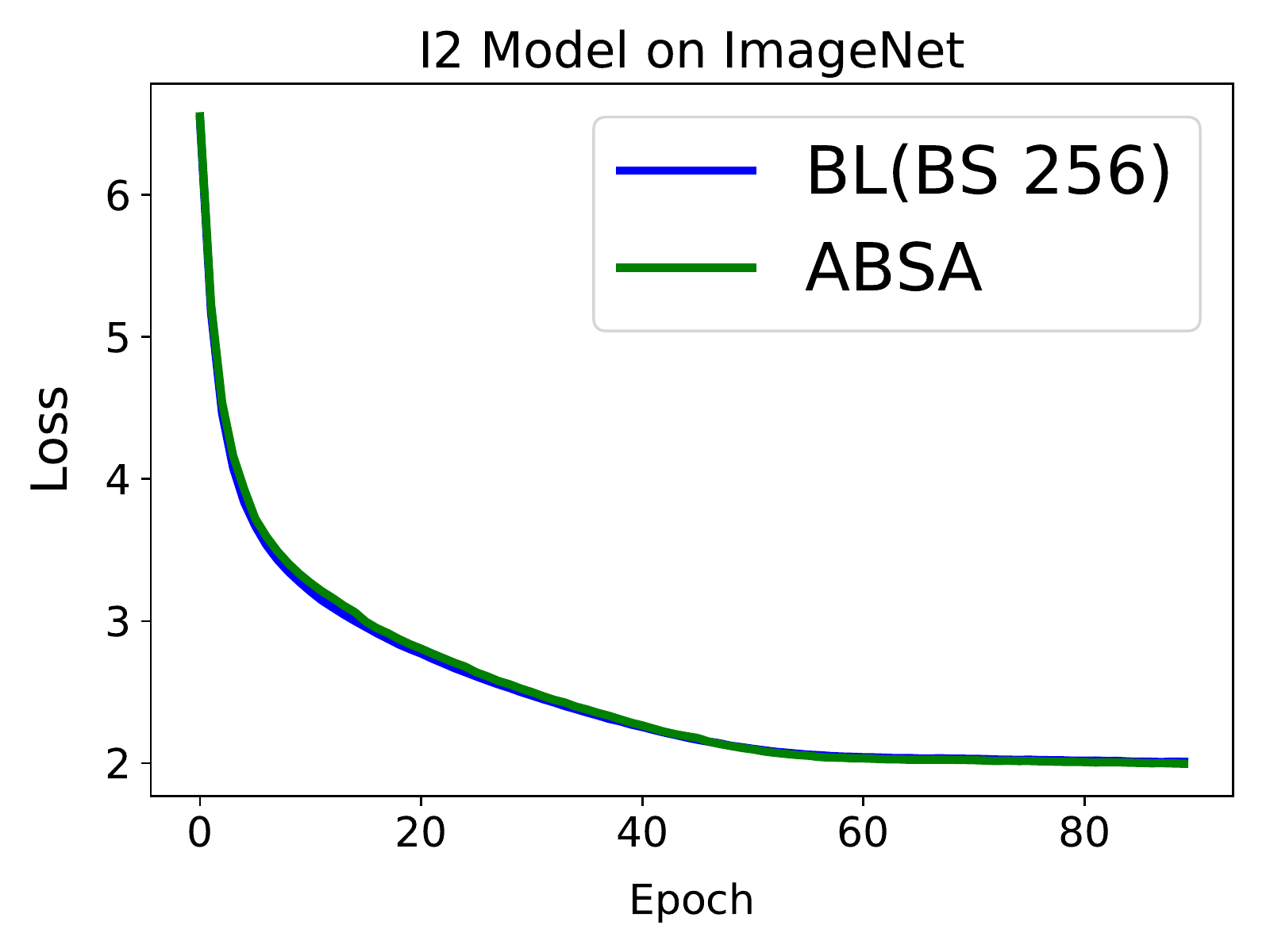}
  \includegraphics[width=.45\textwidth]{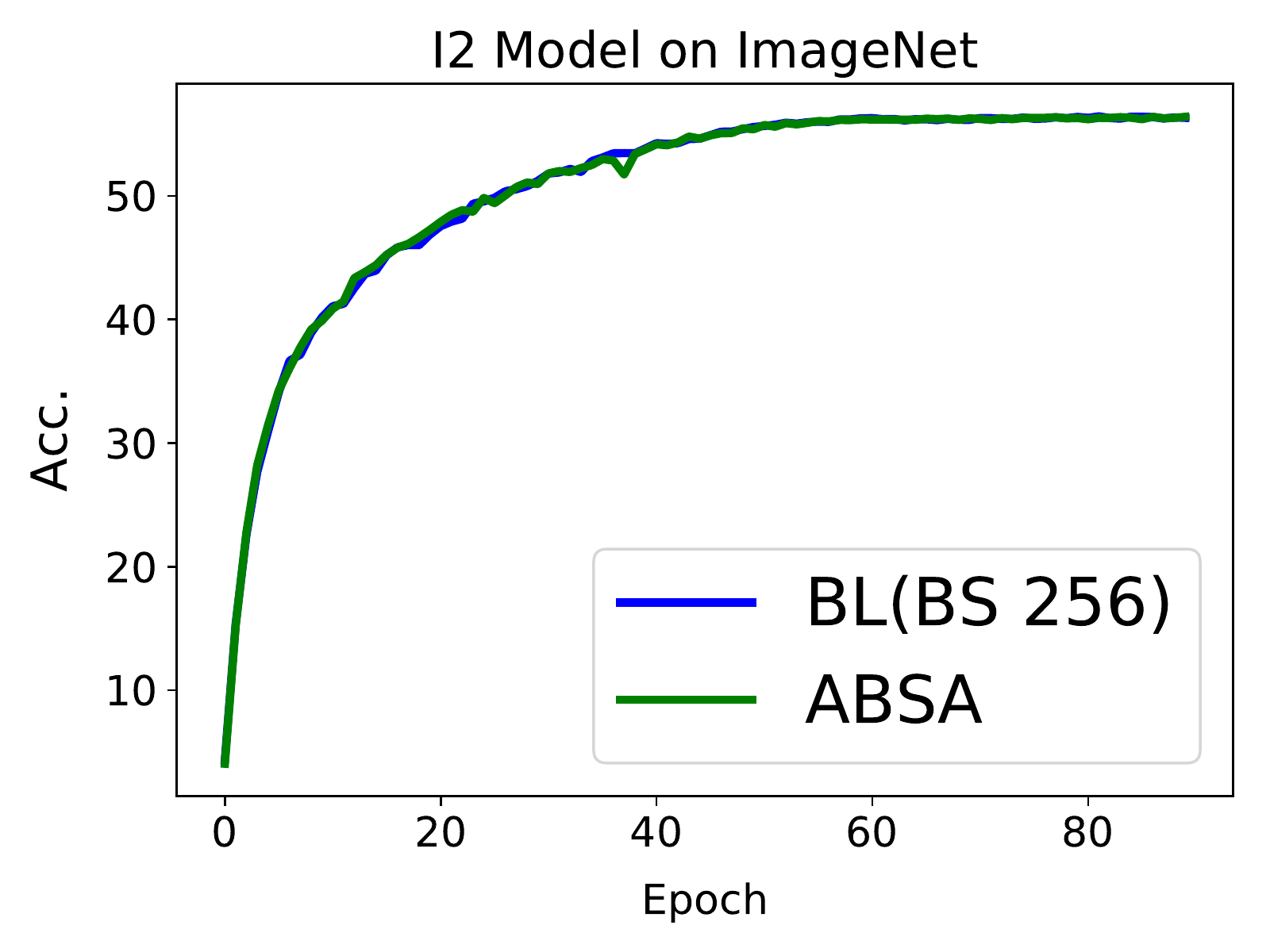}
\end{center}
\caption{I2 model on ImageNet. Training set loss (a), and testing set accuracy (b), evaluated as a function of epochs. 
}
\label{fig:imagenet}
\end{figure}
%%%%%%%%%%%%%%%%%%%%%%%%%%%%%%%%%%%%%%%%%%%%%%%%%%%%%%%%%%%%%%%%%%%%%%%%%%%%%%%%%%%%%%

%%%%%%%%%%%%%%%%%%%%%%%%%%%%%%%%%%%%%%%%%%%%%%%%%%%%%%%%%%%%%%%%%%%%%%%%%%%%%%%%%%%%%%
\begin{figure}[!htbp]
\begin{center}
  \includegraphics[width=.38\textwidth,trim={2.5cm 0 1.5cm 0},clip]{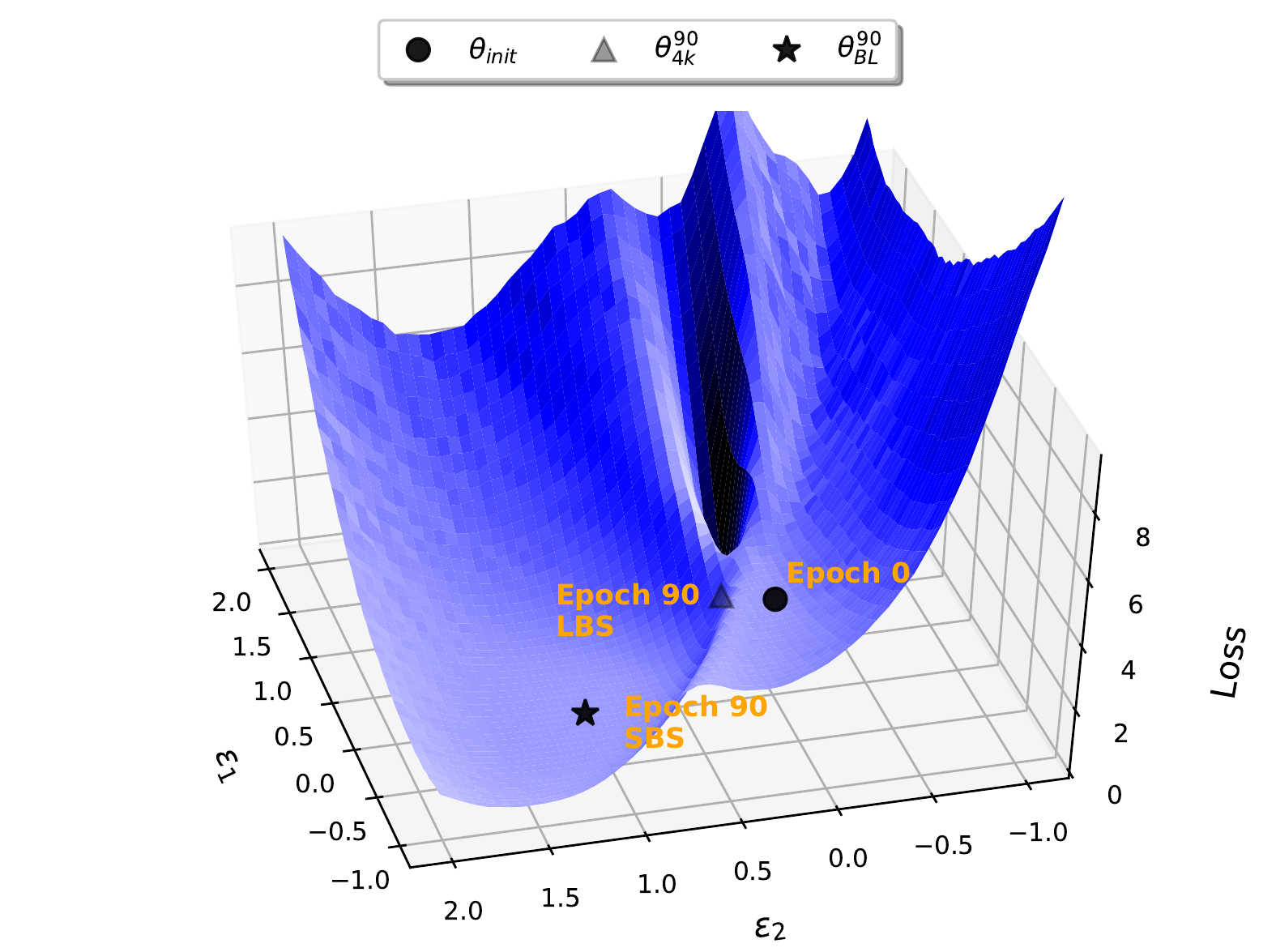}
  \includegraphics[width=.38\textwidth,trim={2.5cm 0 1.5cm 0},clip]{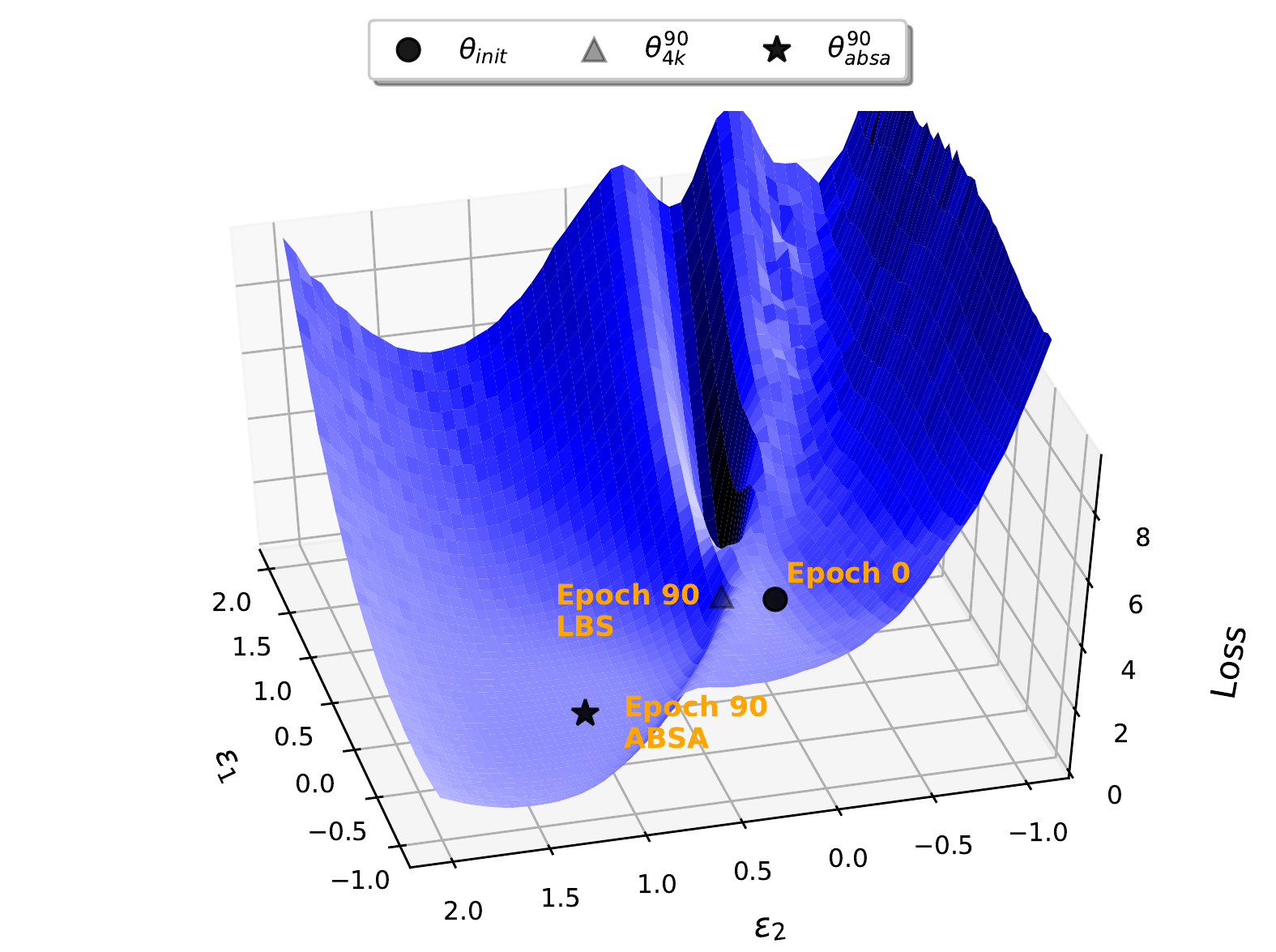}
  \includegraphics[width=.38\textwidth,trim={2.5cm 0 1.5cm 0},clip]{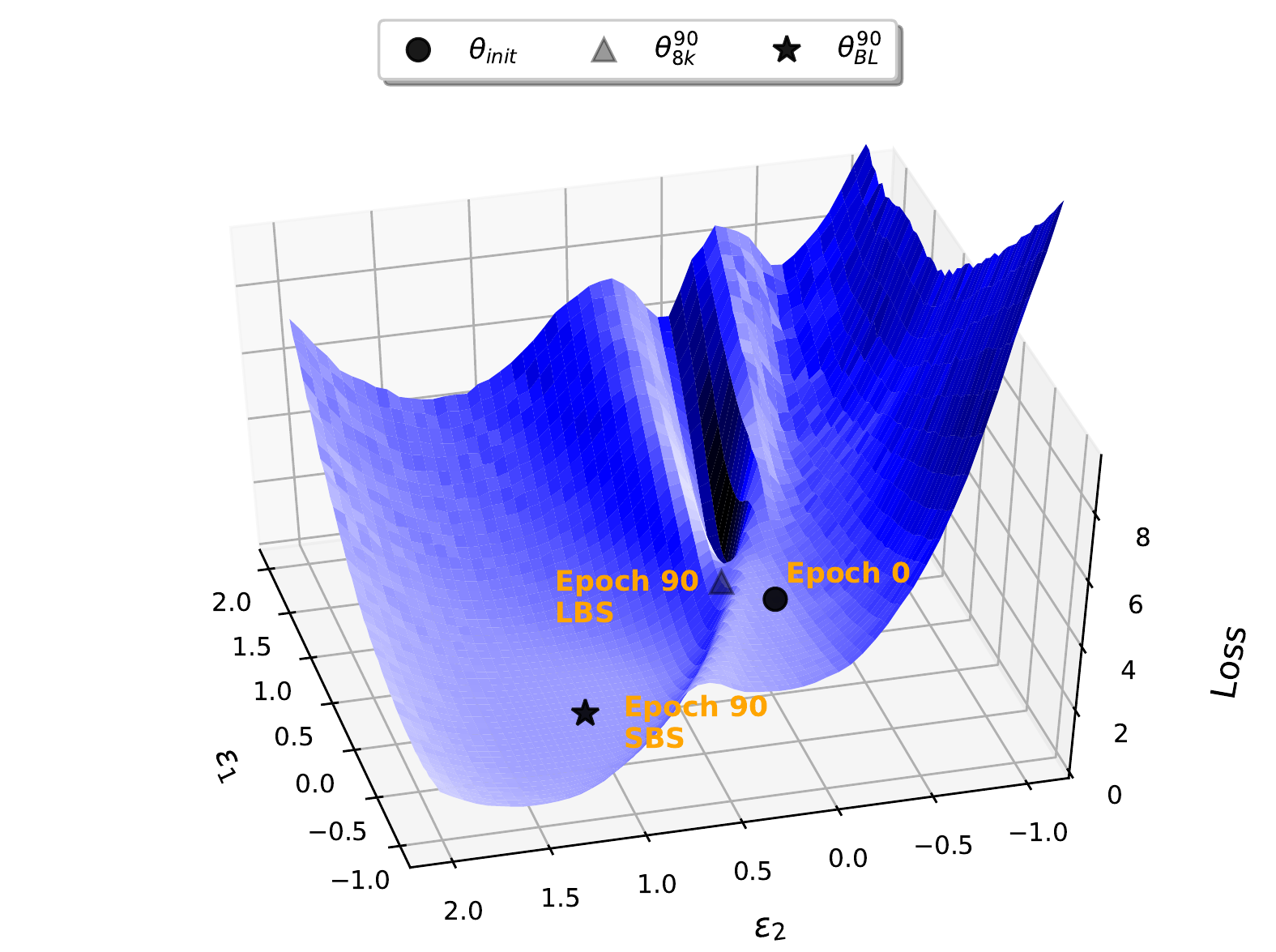}
  \includegraphics[width=.38\textwidth,trim={2.5cm 0 1.5cm 0},clip]{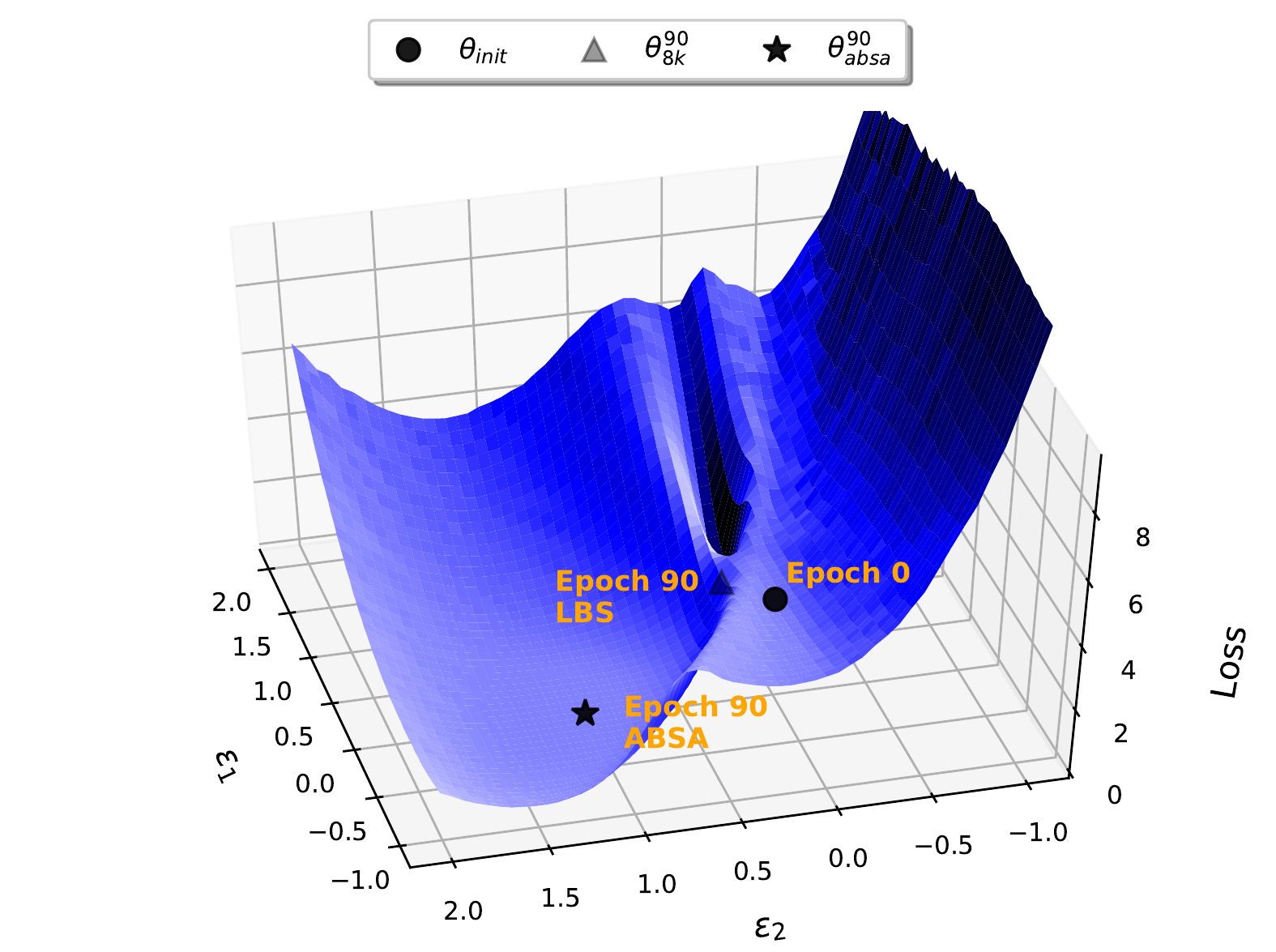}
  \includegraphics[width=.38\textwidth,trim={2.5cm 0 1.5cm 0},clip]{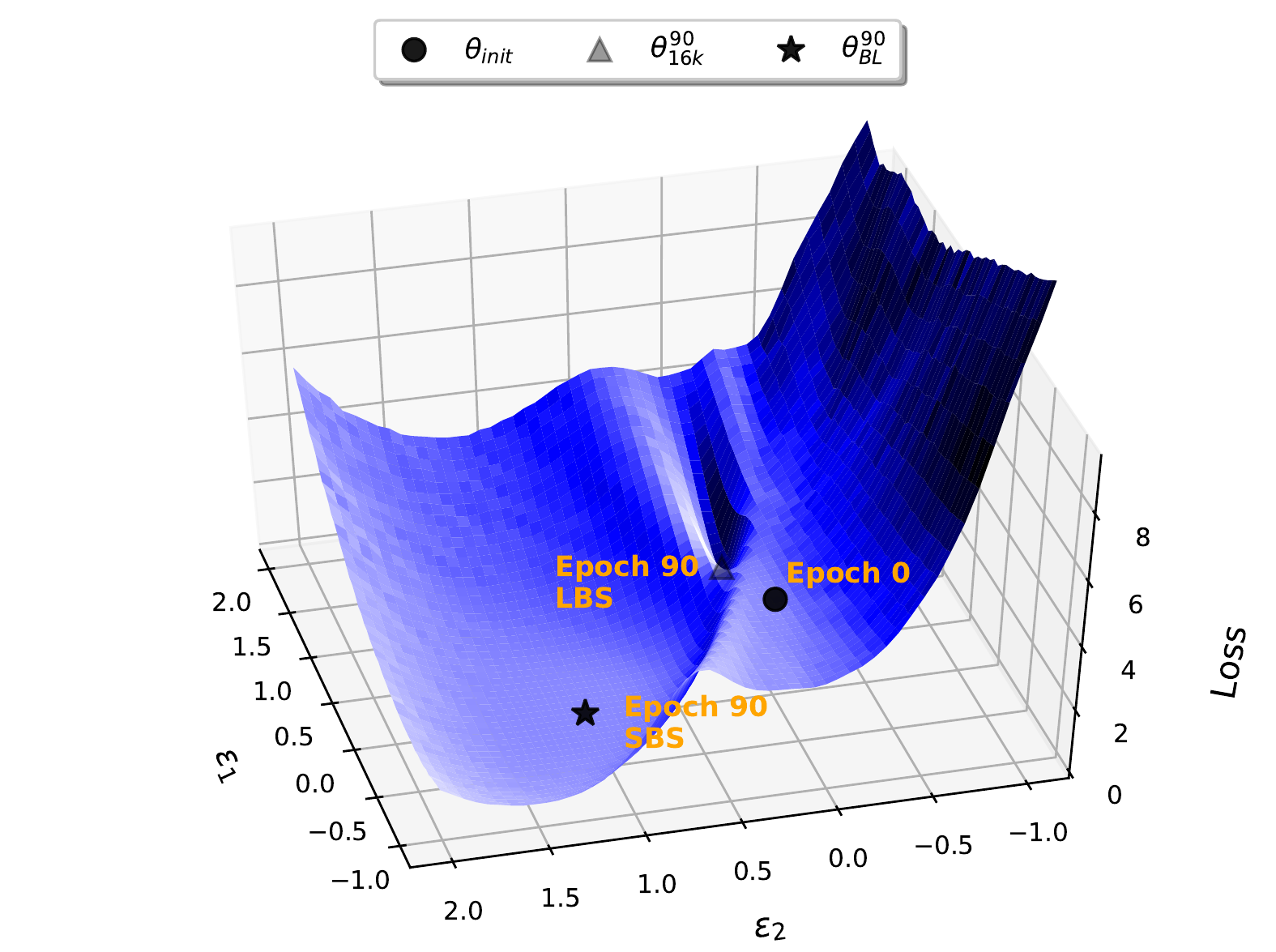}
  \includegraphics[width=.38\textwidth,trim={2.5cm 0 1.5cm 0},clip]{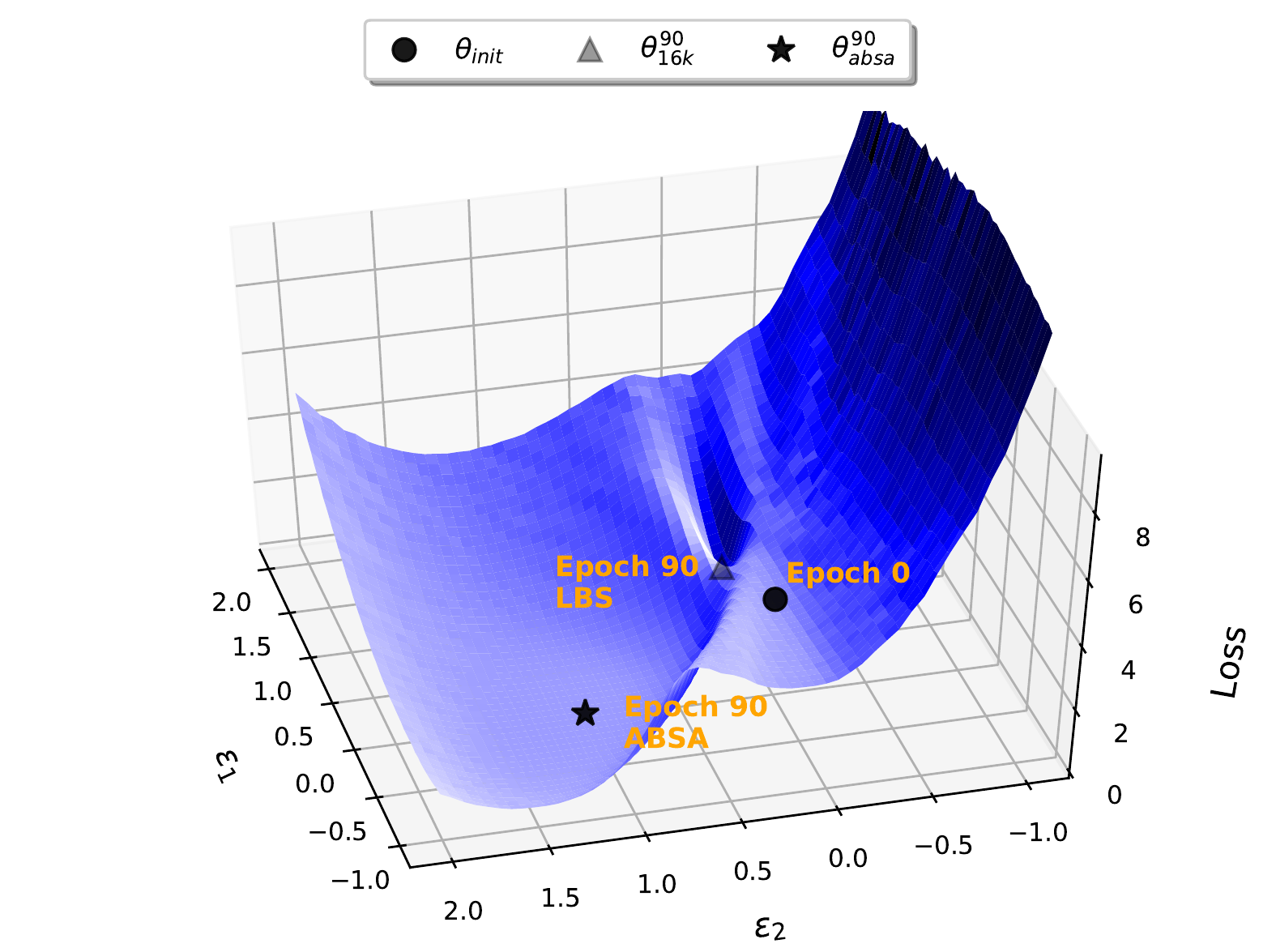}
\end{center}
\caption{(left) 3D parametric plot for C1 model on Cifar-10.
Points are labeled with the number of epochs (e.g. 90) and the technique that was used to arrive at that point (e.g. large batch size, LBS).
The $\epsilon_1$ direction shows
how the loss changes across the path between initial model at epoch 0, and the final model
achieved with Large Batch Size (LBS) of $B=4K,~8K,~16K$. Similarly, $\epsilon_2$ direction computes
loss when the model parameters are interpolated between epoch 0 and final model at epoch 90
with Small Batch Size(SBS). Notice the sharp curvature that Large Batch Size gets attracted to. 
On the right we show a similar plot except that we use ABSA algorithm with final batch of $16K$ rather than SGD with a small batch size for interpolating the $\epsilon_2$ direction. Notice the visual similarity between the point that ABSA converges to after 90 epochs (ABSA, 84.24\% accuracy) and the point that small batch SGD (SBS, 83.05\% accuracy) converges to after 90 epochs. Also note, that both avoid the sharp landscape that large batch gets attracted to sharper landscapes.
}
\label{fig:2d_cifar_resnet_extra_c1}
\end{figure}
%%%%%%%%%%%%%%%%%%%%%%%%%%%%%%%%%%%%%%%%%%%%%%%%%%%%%%%%%%%%%%%%%%%%%%%%%%%%%%%%%%%%%%

%%%%%%%%%%%%%%%%%%%%%%%%%%%%%%%%%%%%%%%%%%%%%%%%%%%%%%%%%%%%%%%%%%%%%%%%%%%%%%%%%%%%%%
\begin{figure}[!htbp]
\begin{center}
    \includegraphics[width=.38\textwidth,trim={2.5cm 0 1.5cm 0},clip]{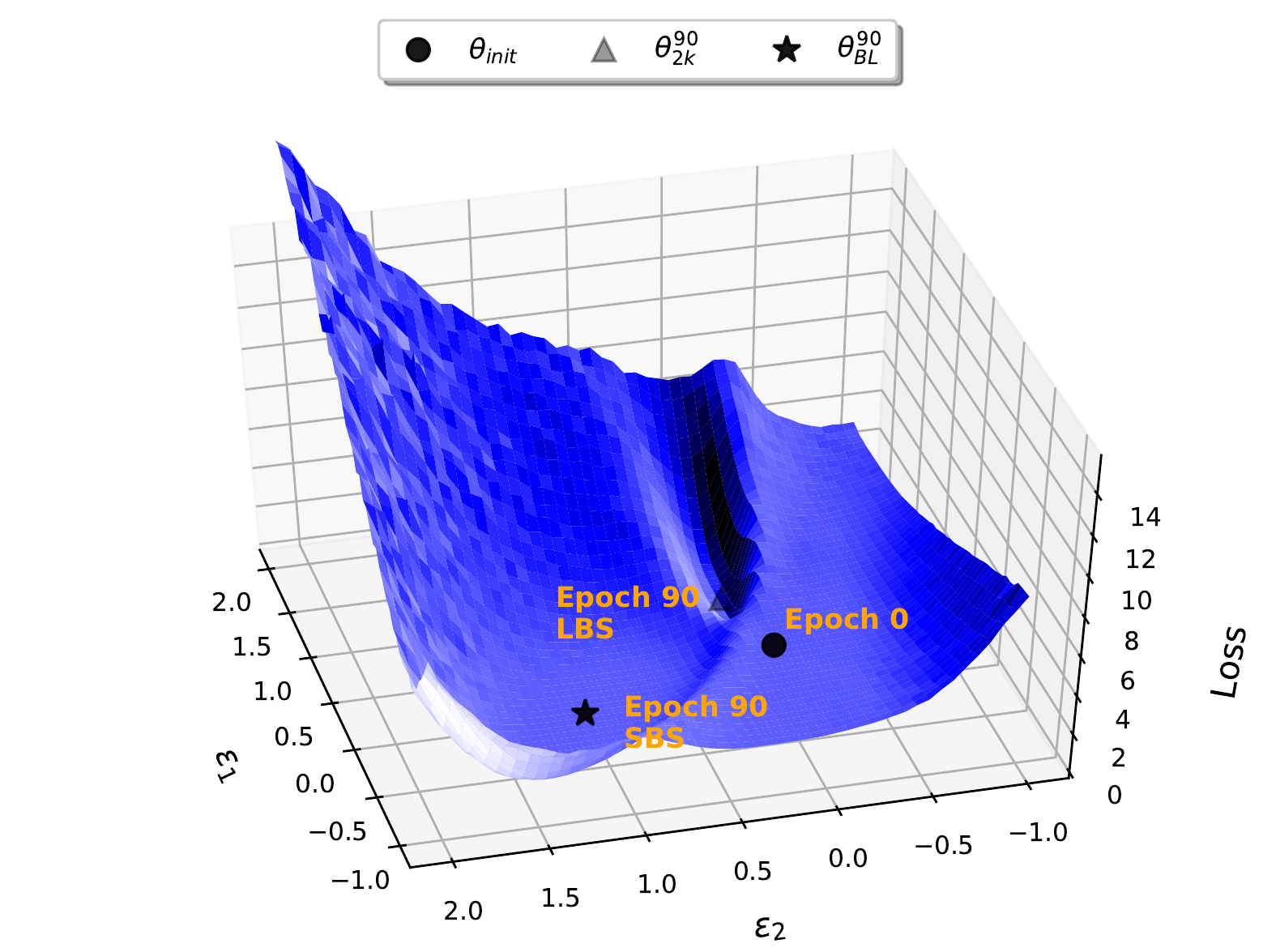}
  \includegraphics[width=.38\textwidth,trim={2.5cm 0 1.5cm 0},clip]{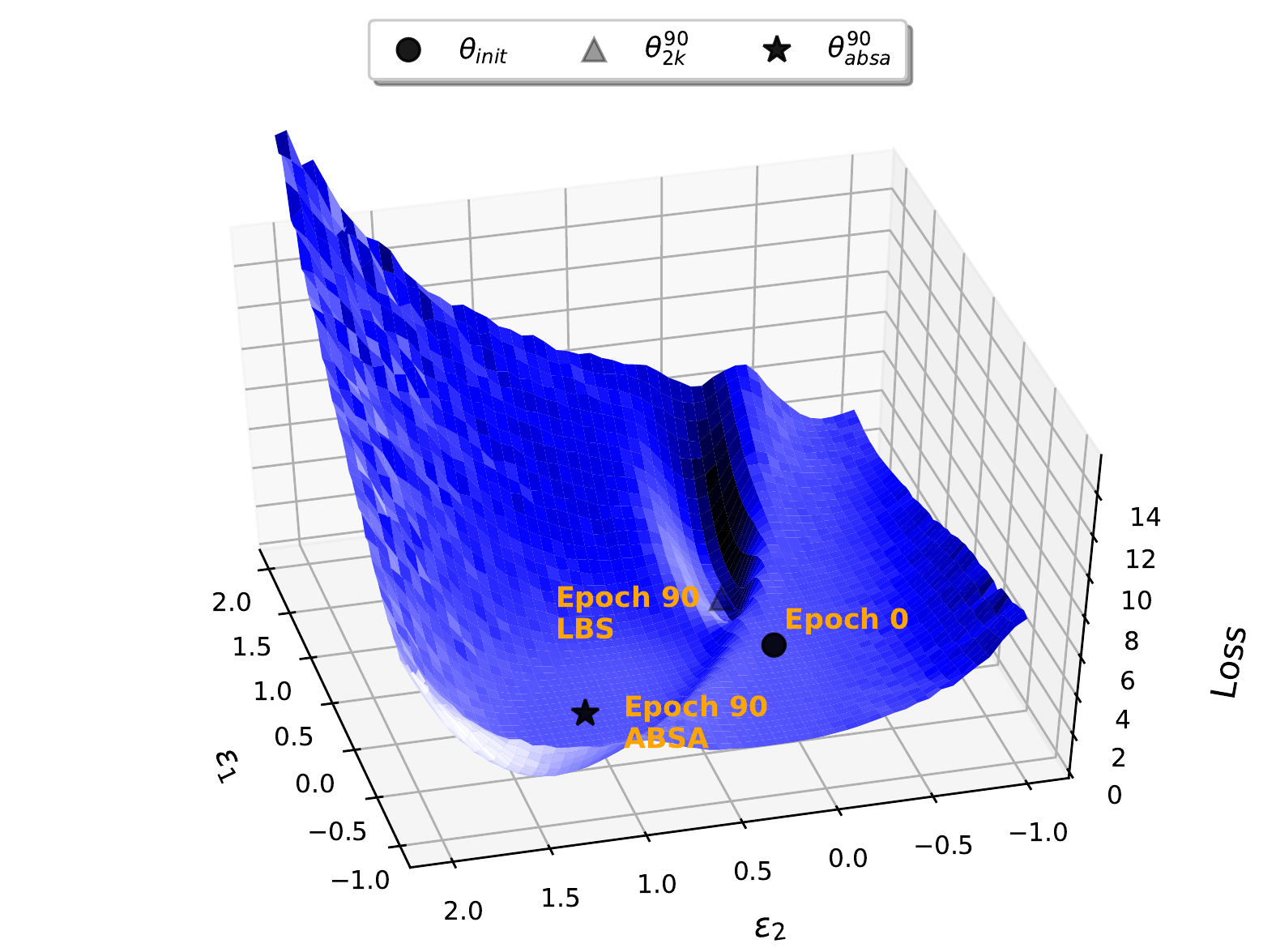}
  \includegraphics[width=.38\textwidth,trim={2.5cm 0 1.5cm 0},clip]{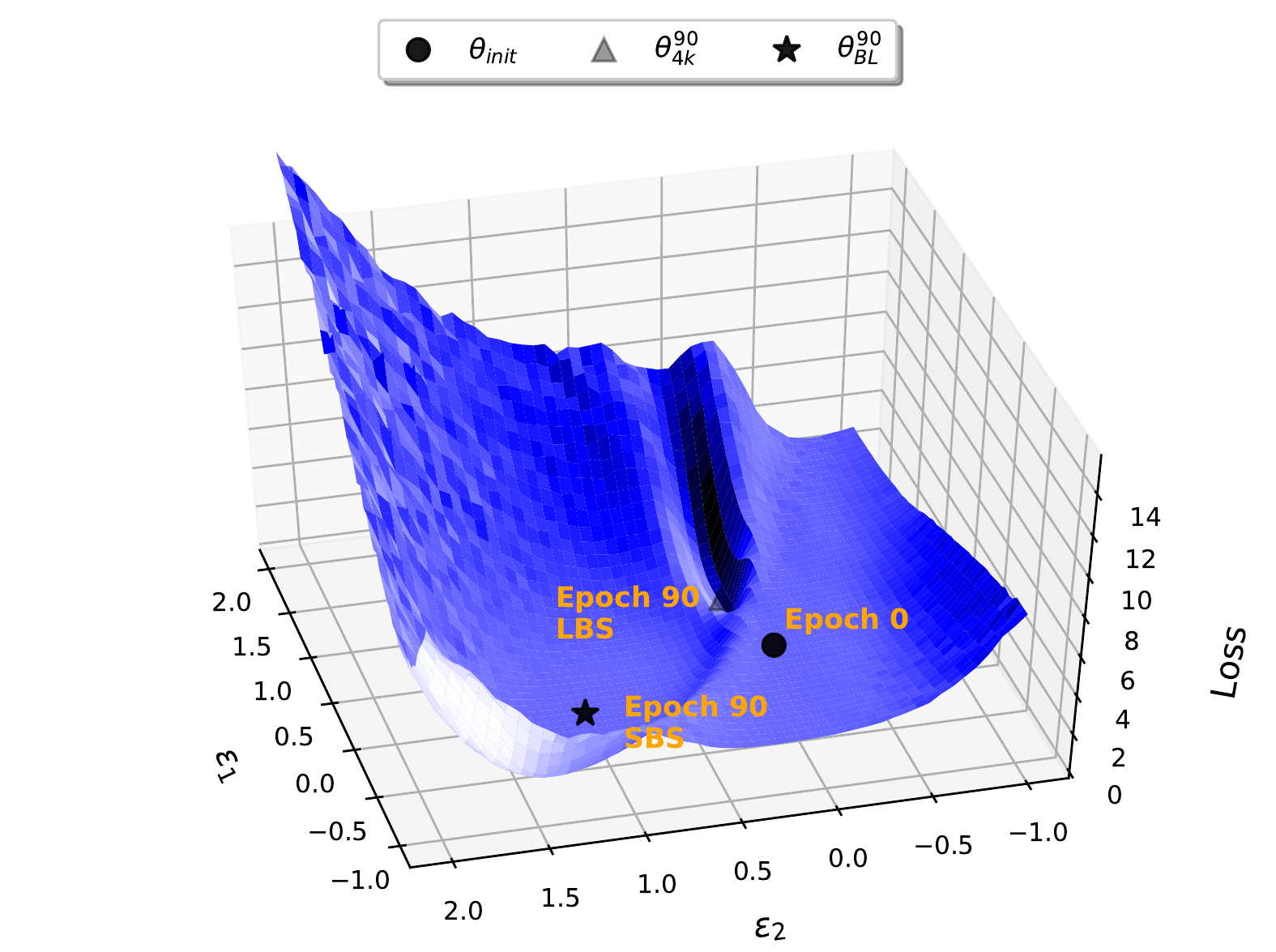}
  \includegraphics[width=.38\textwidth,trim={2.5cm 0 1.5cm 0},clip]{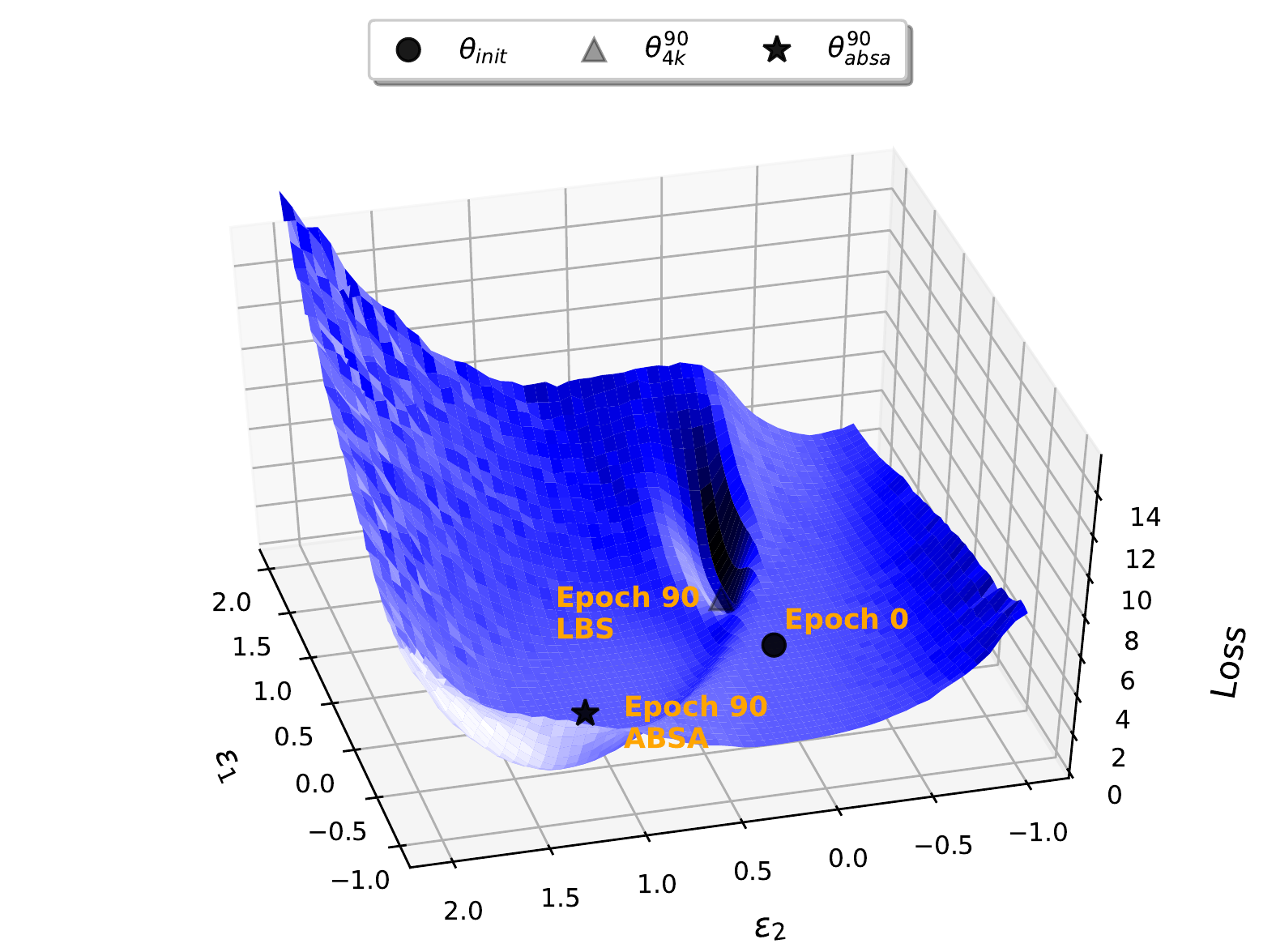}
\end{center}
\caption{(left) 3D parametric plot for C2 model on Cifar-10.
Points are labeled with the number of epochs (e.g. 90) and the technique that was used to arrive at that point (e.g. large batch size, LBS).
The $\epsilon_1$ direction shows
how the loss changes across the path between initial model at epoch 0, and the final model
achieved with Large Batch Size (LBS) of $B=4K,~8K,~16K$. Similarly, $\epsilon_2$ direction computes
loss when the model parameters are interpolated between epoch 0 and final model at epoch 90
with Small Batch Size(SBS). Notice the sharp curvature that Large Batch Size gets attracted to. 
On the right we show a similar plot except that we use ABSA algorithm with final batch of $16K$ rather than SGD with a small batch size for interpolating the $\epsilon_2$ direction. Notice the visual similarity between the point that ABSA converges to after 90 epochs (ABSA, 89.19\% accuracy) and the point that small batch SGD (SBS, 87.64\% accuracy) converges to after 90 epochs. Also note, that both avoid the sharp landscape that large batch gets attracted to sharper landscapes.
}
\label{fig:2d_cifar_resnet_extra_c2}
\end{figure}
%%%%%%%%%%%%%%%%%%%%%%%%%%%%%%%%%%%%%%%%%%%%%%%%%%%%%%%%%%%%%%%%%%%%%%%%%%%%%%%%%%%%%%

%%%%%%%%%%%%%%%%%%%%%%%%%%%%%%%%%%%%%%%%%%%%%%%%%%%%%%%%%%%%%%%%%%%%%%%%%%%%%%%%%%%%%%
\begin{figure}[!htbp]
\begin{center}
  \includegraphics[width=.9\textwidth]{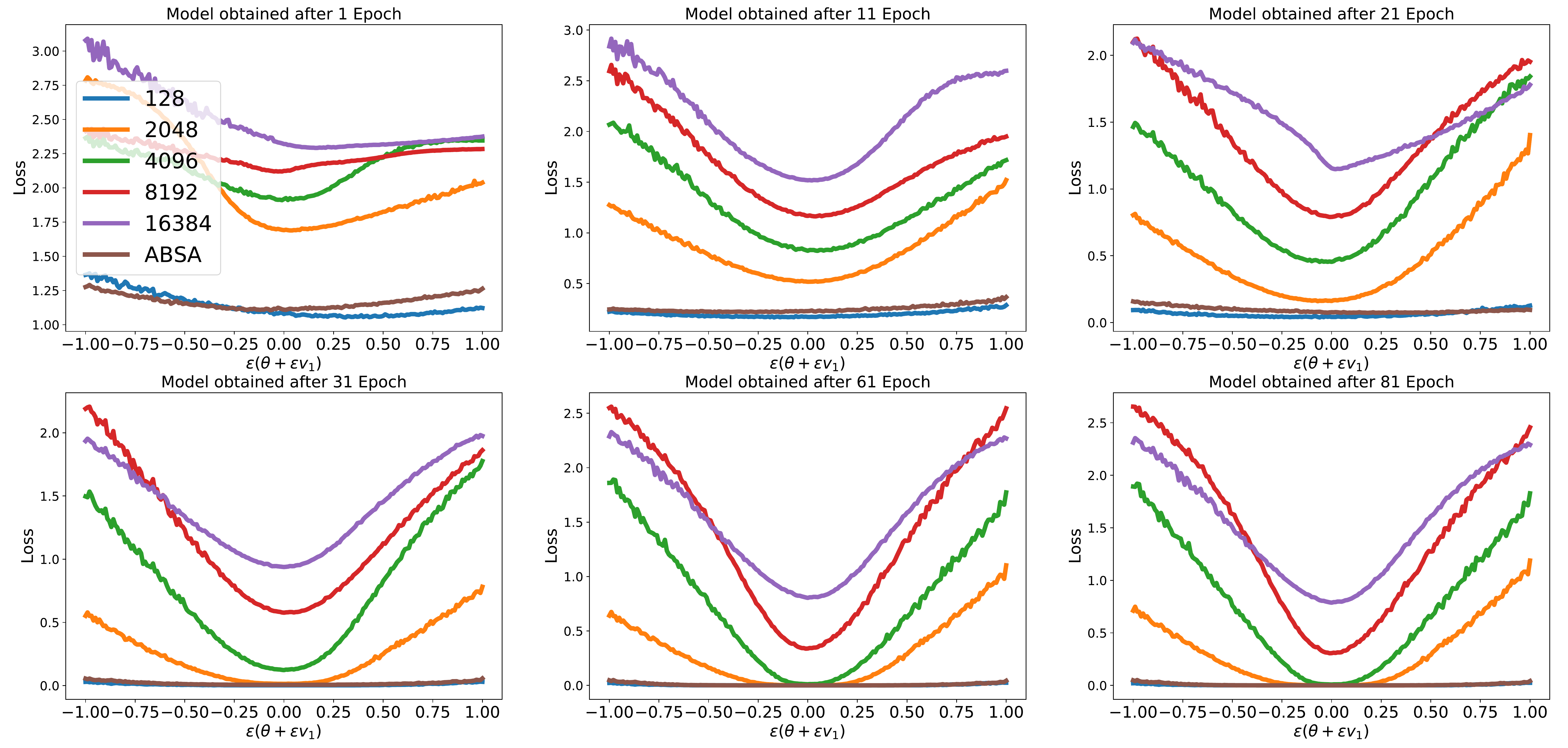}
\end{center}
\caption{The landscape of the loss is shown when the model parameters
are perturbed along the dominant Hessian eigenvector, $v_1$, for C1 on Cifar-10 dataset. Here $\epsilon$ is a scalar that perturbs the model parameters along $v_1$. 
The top rows show the landscape for epochs 1, 11, and 21 and the bottom row shows
epochs 31, 61, and 81. One can clearly see, that larger batches continuously get stuck/attracted
in areas with larger curvature whereas small batch SGD or ABSA (with batch of 16K) can escape such landscape.
For detailed generalization error please see~\tref{tab:abs_cifar10_resnet18}.
}
\label{fig:loss_landscape_1d_resnet18}
\end{figure}

% %%%%%%%%%%%%%%%%%%%%%%%%%%%%%%%%%%%%%%%%%%%%%%%%%%%%%%%%%%%%%%%%%%%%%%%%%%%%%%%
% %%%%%%%%%%%%%%%%%%%%%%%%%%%%%%%%%%%%%%%%%%%%%%%%%%%%%%%%%%%%%%%%%%%%%%%%%%%%%%%

\end{document}